\documentclass[journal,twoside]{IEEEtran}
\IEEEoverridecommandlockouts
\usepackage{cite}
\usepackage{amsmath,amssymb,amsfonts}
\usepackage{algorithmic}
\usepackage{balance}
\usepackage{graphicx}
\usepackage{multirow}
\usepackage{textcomp}
\usepackage{csquotes}
\usepackage{framed}
\usepackage{xcolor}
\usepackage{siunitx}
\def\BibTeX{{\rm B\kern-.05em{\sc i\kern-.025em b}\kern-.08em
    T\kern-.1667em\lower.7ex\hbox{E}\kern-.125emX}}
\usepackage{color}
\definecolor{dkgreen}{rgb}{0,0.6,0}
\definecolor{gray}{rgb}{0.5,0.5,0.5}
\definecolor{mauve}{rgb}{0.58,0,0.82}
\usepackage[procnames]{listings}

\begin{document}

\title{Physics-Guided Adversarial Machine Learning for Aircraft Systems Simulation}

\author{Houssem~Ben~Braiek,
        Thomas~Reid,
        and~Foutse~Khomh,~\IEEEmembership{Senior Member,~IEEE,}
\thanks{H. Ben Braiek and F. Khomh are with the Department
of Computer and Software Engineering, Polytechnique Montr\'{e}al, Quebec, H3T 1J4 Canada (e-mail: houssem.ben-braiek@polymtl.ca; foutse.khomh@polymtl.ca).}
\thanks{T.Reid is with Bombardier Aerospace Inc., Dorval, Quebec, H9P 1A2 Canada (e-mail: thomas.x.reid@aero.bombardier.com).}
\thanks{Manuscript received August 25, 2021; revised August 25, 2021.}}
\markboth{IEEE TRANSACTIONS ON RELIABILITY,~Vol.~x, No.~x, August~2021}{Ben Braiek \MakeLowercase{\textit{et al.}}: Physics-Guided Adversarial Machine Learning for Aircraft Systems Simulation}
\maketitle
\begin{abstract}
In the context of aircraft system performance assessment, deep learning technologies allow to quickly infer models from experimental measurements, with less detailed system knowledge than usually required by physics-based modelling. However, this inexpensive model development also comes with new challenges regarding model trustworthiness. This work presents a novel approach, physics-guided adversarial machine learning (ML), that improves the confidence over the physics consistency of the model. The approach performs, first, a physics-guided adversarial testing phase to search for test inputs revealing behavioral system inconsistencies, while still falling within the range of foreseeable operational conditions. Then, it proceeds with a physics-informed adversarial training to teach the model the system-related physics domain foreknowledge through iteratively reducing the unwanted output deviations on the previously-uncovered counterexamples. Empirical evaluation on two aircraft system performance models shows the effectiveness of our adversarial ML approach in exposing physical inconsistencies of both models and in improving their propensity to be consistent with physics domain knowledge.
\end{abstract}


\section{Introduction}
\IEEEPARstart{C}{ivil} aircraft product development involves complex feedback loops of development and optimization to meet certification and performance requirements. This highly-iterative, complex development workflow requires a combination of extensive domain knowledge, advanced simulation technology, expanded engineering experience and flight test control in order to build high confidence over the designed systems’ behaviors. Engineers heavily rely on computerized design aid solutions to model the physical system and assess its conformance with desired requirements regarding principal functionality, safety and reliability. Thus, the system model must be qualified to appropriately reproduce the system's behavior throughout the range of foreseeable operational conditions. This is utterly important to avoid uncovering issues late during production or in-flight testing. Traditionally, physics modelling is used to analyze the system-related physics principles and \textit{apriori} knowledge of system design, in order to develop physics-based models with detailed representations of underlying physics processes and strong priors stemming from first principles and governing equations. Despite the gain in trustworthiness resulting from the structural understanding of physical models’ latent variables and equations, they often require comprehensive information about the aircraft physics, and rely on challenging implementation, calibration, verification and validation processes~\cite{marini2002verification, roache1998verification}. Recently, computational physics and system engineering researchers~\cite{ioannou2018structural, ren2018learning, stewart2017label, muralidhar2019physics} have been increasingly exploring the potential of using machine learning (ML) to alleviate the cost of hand-crafting physical system models. The current surge of ML field, especially modern deep learning (DL)~\cite{wang2020recent}, brought forward a wide array of model architectures and training techniques that are extremely powerful at approximating any mapping function between variables based on observations and measurements data, without prior knowledge of the underlying governing processes. Nevertheless, the inexpensive implementation cost of DL-based systems comes with significant challenges regarding system trustworthiness as evidenced by the emergence of adversarial machine learning~\cite{biggio2018wild}. First, the experimental data cannot be an appropriate substitute for specifications in the absence of evidence supporting that the data distribution matches the operational conditions. Thus, there is no guarantee that a deep neural network (DNN) trained over a finite set of experimented operational scenarios, could generalize to behave correctly for new operational conditions that were not represented in the original training datasets. Second, the black-box nature of modern learning models and their resulting performance-driven complex architectures no longer allows a full understanding of the complete structural design. Hence, there is no direct method to assess the relevance of the learned latent patterns, whether they support generalizability, or they result from a coincidental optimality of memorizing some spurious and noisy patterns in the data. Below, we describe the contribution of this paper:

First, the high-level physics specifications including first principles and \textit{apriori} system design properties can be expressed in the format of input-output sensitivity relations. Then, we design a physics-guided adversarial testing method that assesses the conformance of the ML model with the system physics-grounded requirements through validating that the model under test predicts consistent outputs for all the neighborhood local input space around each original data. The effectiveness evaluation of our proposed testing approach was conducted on two industrial cases of aircraft system performance models. Results show that our physics-guided adversarial testing method succeeds in exposing physics inconsistencies in the predictions of the studied neural networks, and our swarm-based search algorithm outperformed by far genetic algorithm and random sampling baseline in terms of exposed physics inconsistencies counts.

Next, we design a physics-informed adversarial training technique that leverages the counterexamples found by the adversarial testing method to guide the parameters optimization routines towards finding the best-fitted model with lowest amount of deviation errors and physics inconsistencies in its outputs. Evaluation of our approach as fine-tuning phase on the previously-tested system performance models shows that our physics-informed adversarial training can improve the model's propensity to be consistent with the specified high-level system requirements. Furthermore, we show how the proposed physics-informed fine tuning phase can improve the model performance depending on the quality of the exposed adversarial inputs. In our case, GA-enabled search algorithm was capable of generating few adversarial examples, but relevant for the model's generalizability improvement.

The remainder of this paper is organised as follows.\\ Section~\ref{sec:related_work} summarizes the relevant literature and introduces the adversarial ML and software testing concepts adapted by our approach. Section~\ref{sec:adv_testing} presents our novel physics-guided adversarial testing for ML models. Section~\ref{sec:adv_training} describes a designed physics-informed loss to improve the physics consistency of the on-training ML solution. Section~\ref{sec:evaluation} reports evaluation results, while Section~\ref{sec:conclusion} concludes the paper.

\section{Background and Related Works}
\label{sec:related_work}
This section describes the essential concepts of adversarial machine learning and search-based software testing and review the relevant literature.
\subsection{Adversarial Testing}
In the context of computer vision, Szegedy et al.~\cite{szegedy2013intriguing} show that universal approximator and high learning capacity neural networks such as deep convolutional neural networks can react in unexpected and incorrect ways to even slight pixel-based perturbations of their benign inputs. Such finding leads researchers to invent systematic adversary attacks to stress these vulnerabilities in the model learned patterns. We refer to the extensive survey by Biggio and Roli~\cite{biggio2018wild} for more details about adversarial machine learning. In the following, we introduce the analytical formulation of the adversarial testing problem. Indeed, the objective of adversarial testing is to design adversary that allows us to detect the data variation ($\delta$) and the natural input ($x$) yielding an adversarial example $\hat{x} = x + \delta$, for which the model does not satisfy a given property $C$. Then, the adversary can be leveraged to improve the satisfiability of the studied property by the model over foreseen inputs. The model robustness represents the first major property that has been relatively well-studied for supervised learning problems using adversarial testing. For classification problem, we say the model $f$ is delta-local robust at the input $x$ if for any $\hat{x}$: $\quad \lVert x - \hat{x} \rVert \leq \delta \longrightarrow f(x) = f(\hat{x})$
$\lVert.\rVert_p$ represents $p$-norm for distance measurement. The commonly used $p$ cases in machine learning testing are $2$, $\infty$. For regression problems, an epsilon $\epsilon$ is introduced to transform the equality between the resulting discrete outputs into the below-mentioned inequality to support robust comparisons between continuous outputs: $\forall \hat{x},\quad \lVert x - \hat{x} \rVert \leq \delta \longrightarrow |f(x) - f(\hat{x})|\leq\epsilon$.

Robustness against adversarial examples raises the importance of building adversarial attacks and defense techniques aiming at detecting and immuning models against these vulnerabilities early on. While most research studies have focused on local robustness adversarial testing in the context of computer vision system, our present work study the extension of this concept to support other interesting properties: (1) \textit{Model physics consistency} refers to verifying that the predictions are consistent with the physics domain knowledge, which combines first principles and \textit{apriori} system design knowledge; (2) \textit{Model relevance} represents the adequacy of the learning capacity to be less sensitive to overfitting and more robust to fit new samples from the data distribution. 
\subsection{Adversarial Training}
\label{adv_train}
Adversarial training is one of the most notable countermeasure, which was first proposed by Goodfellow et al.~\cite{goodfellow2014explaining}, to improve the robustness of DNNs against adversaries, by re-training the model on the adversarial examples found. In practice, the label of original input can be assigned to its perturbed adversarial versions, then, the latter are inserted into the training data to be distributed over the batches during the learning updates. For regression problem, Nguyen and Raff~\cite{nguyen2018adversarial} add regularization term in the loss function to encourage the numerical stability around delta-local neighborhood. Given $\lVert\Delta x\rVert_p < \epsilon$ to denote that we are sampling a point $\Delta x$ uniformly from the $p$-norm ball of radius $\epsilon$, we define adversarial robust loss as follows: 
\begin{equation}
l(y, f(x)) + \lambda\times\mathop{\mathbb{E}}_{\Delta x:\lVert\Delta x\rVert_p < \epsilon}[l(f(x), f(x+\Delta x)]
\end{equation}
where $\lambda \in \mathbb{R}_+$ controls the strength of defense regularization penalty.
In fact, the above-mentioned adversarial loss is composed of two components: the original loss measures the difference between the target output $y$ and the predicted output $\hat{y} = f(x)$, and the regularization penalty imposes that the expectation of the output between a point $x$ and all points within an $L_p$ ball with radius $\epsilon$ around $x$ are the same. Regarding adversarial training, our designed physics-informed adversarial training places more generic cost in the loss function too discourage converging to undesirable input-output mappings, which might be totally incoherent with underlying physics domain foreknowledge.
\subsection{Search-based Software Testing}
\label{SBST}
In the present work, we use Search-based Software Testing (SBST)~\cite{mcminn2004search}, which leverages metaheuristics optimizing search technique to automatically generate test inputs. It is a test generation technique that has been widely used for traditional software testing. Recently, Ben Braiek and Khomh~\cite{deepevolution} highlighted the potential of SBST for testing DL software systems. Nature-inspired population-based metaheuristics possess intrinsically complex routines and non-determinism that make them high potential candidate for spotting vulnerable regions in the large, multi-dimensional input space of the DL models. In the following, we describe two widely-used metaheuristic algorithms in testing machine learning software systems. Indeed, they succeed in crafting black-box adversarial examples efficiently with few queries and they have achieved white-box comparable performances in different application domains including computer-vision~\cite{bhambri2019survey, mosli2019they, alzantot2019genattack, chen2019poba}, natural language processing~\cite{wang2019natural,zang2020word, du2020hybrid}, and speech assistance DNNs~\cite{du2020sirenattack}.\\

\textbf{Particle Swarm Optimization(PSO)~\cite{PSO}} mimics the behavior of a swarm of birds to search a very large space of candidate solutions. PSO maintains a population of candidate solutions called particles. As particles move in the search space, they seek better solutions to the problem based on their fitness values. The inertia weight, \textit{w}, affects particle velocity and search space expansion. The movement of each particle within an iteration is guided by its local best position, $p_{best}$, which refers to exploring its best neighbour regions, while at the same time being guided toward a global best position¸ $g_{best}$, by all particles, which refers to exploiting the highest fitness region found. The PSO algorithm employs two trust coefficients, $\varphi_p$ and $\varphi_g$, which set up a particle's confidence in itself (cognitive coefficient) and in its neighbors (social coefficient).\\

\textbf{Genetic Algorithm(GA)~\cite{GA}} emulates the behavior of biological evolution, including basic selection, crossover, and mutation operations that can lead to, potentially, better individuals in every generation. By using a tournament selection strategy, GA selects a $r_{parents}$ of individuals to become parents for the next generation. Breeding is done by randomly picking matching pairs of parents and using one of these binary crossovers~\cite{picek2010comparison}, one-point, two-point, or uniform, to produce new offspring. Each parent in a couple will be assigned a relative importance based on its fitness level. At the end of the breeding, we apply mixed random mutations to alter the features of the offspring in order to maintain and introduce diversity into the new generation. Indeed, every descendant can be subject to a mutation, depending on a fixed probability, $p_{mutation}$.

To the best of our knowledge, existing SBST-based adversarial input generators focus on assessing the robustness of the ML model against slightly-perturbed versions of the original inputs, however, adversarial examples are often criticized for being non-representative of natural input data. Compared to such works, the novelty of our SBST-based approach is that it leverages the flexibility of gradient-free metaheuristic algorithms to codify physics-grounded rules and constraints in order to search for adversarial examples satisfying foreseeable data constraints but violating the system-related physics domain specifications. This allows to assess the physics consistency of the model and enhances confidence in its fitness to the application-specific requirements. 
\subsection{Physics-guided Machine Learning}
The impressive efficiency of deep learning in solving industrial problems gave rise to several researches that worked on increasing the performance of data-driven system models using domain knowledge. In~\cite{ren2018learning,stewart2017label}, a custom loss was proposed based on the domain knowledge, which eliminates entirely the need for supervision data. Karpatne et al.~\cite{karpatne2017theory} propose to integrate a domain knowledge regularizer in neural networks to influence the model optimization towards better generalization performance. Apart from modifying the loss optimization problem, Ioannou et al.~\cite{ioannou2018structural} shows that \textit{apriori} structural knowledge could be included into the model architecture design. In a similar fashion, researchers leverage prior knowledge about the problem to incorporate feature invariance~\cite{ling2016reynolds}, to enable representations consistent with physics domain knowledge through either implicit physics rules~\cite{seo2019differentiable} or explicit physics-based constraints~\cite{leibo2017view}. 
None of these former researches can be directly applicable to encode the physical relationships required in modelling our target aircraft systems simulators. Instead, our physics-guided adversarial ML approach enables more flexibility as it is designed and implemented independently from both of the deployed DL technologies and the simulated physics-intensive system's details. Indeed, our physics-guided adversarial testing accepts system-related physics domain knowledge in the format of input-output sensitivity relations with an expected tolerance. Then, it generates optimized test cases to validate, in a black-box fashion, the consistency of the model's input-output mappings that have been statistically inferred from the data. This is also applicable for the proposed physics-informed regularization that performs a data driven fine-tuning of the model relying on the adversarial inputs revealed during the test, which have exposed the model's violations in regards to the specified physics-grounded sensitivity rules.
\section{Physics-guided Adversarial Testing}
\label{sec:adv_testing}
We have developed an approach to assess how well a model is consistent with the foreknown system-related physics domain knowledge. In the following, we detail the three principles of the proposed approach, its integrated components, and the resulting physics-guided adversarial testing workflow.  
\begin{figure*}[h]
\centering
\includegraphics[scale=0.65]{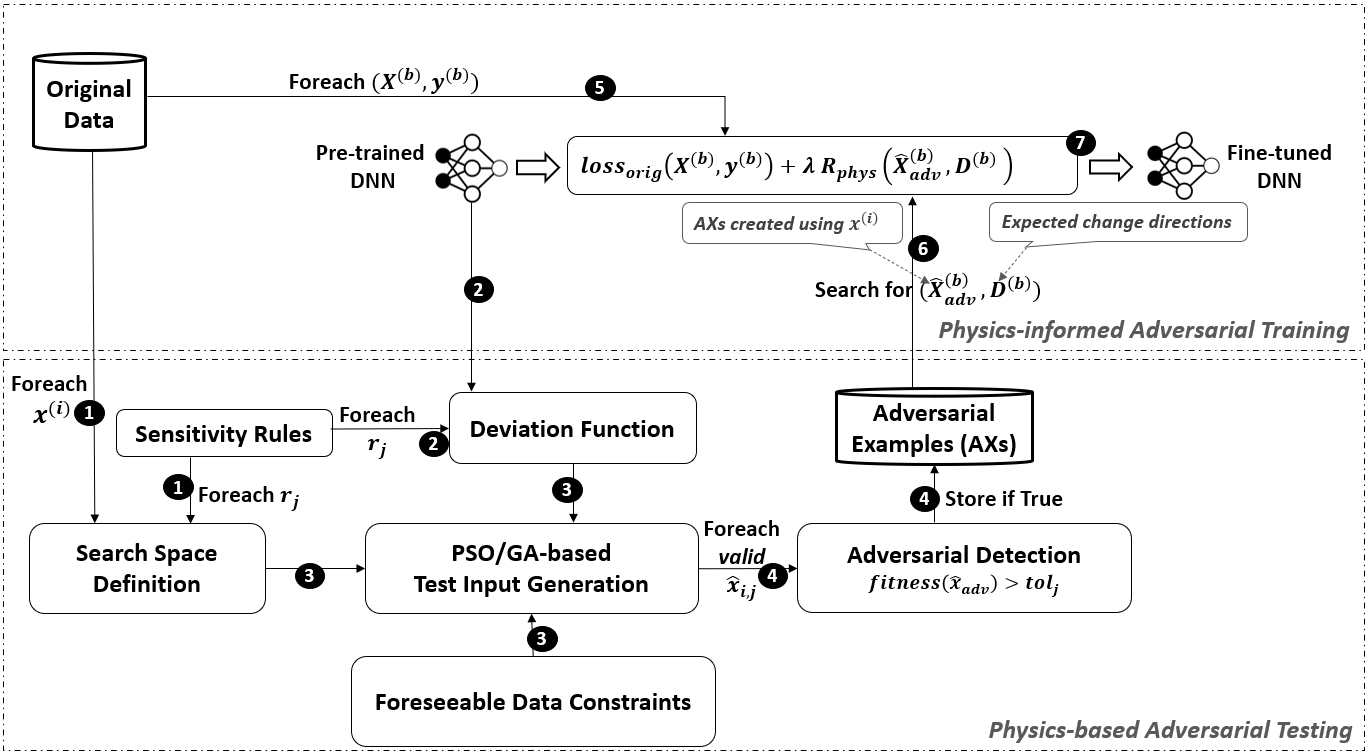}
\caption{Overview of Physics-Guided Adversarial Machine Learning Phases and Workflow}
\label{fig:overview_phys_adv_learning}
\end{figure*}
\subsection{Specification of Physics Domain Knowledge}
Aircraft product development heavily relies on modelling physical processes and phenomena to enable the interpretation of the interactions between the input quantities, $x$, and dependent variables, $y$ that are observed during a flight. In contrast, data-driven modelling consists of inferring statistically the mapping between the input quantities and the observed variables from the collected experimental data without prior knowledge of the underlying governing physics phenomena. Both system modelling approaches would serve similar engineering use cases that mainly require the assessment of system behavior at different operating conditions (i.e., normal and extreme flight scenarios). Even if the model's mapping function does not include explicitly physics equations, the statistically-inferred mappings should respect the system-related physics processes as well. Hence, we propose the specification of the system's desired properties in terms of must-hold relationships between the input sensors data and the target quantity of interest. Mathematically, the system domain expert codify these foreknown relationships in a format of sensitivity rules that map a subset of oriented input features variations (i.e., signed value changes) to anticipated variation trend of the output. For instance, we might foreknow that increase of $x_1\nearrow$ and/or decrease of $x_2\searrow$ lead to the increase of $f(x_1,x_2) = y\nearrow$. These sensitivity rules are designed to be applied locally over the genuine data points when system engineering experts can confirm that combined input features variations usually map to higher order effects that are not necessarily detectable locally within predefined bounds. Furthermore, we highlight that our defintion of sensitivities include the input features variations leading to no trend conclusion (i.e., a constant model output). These special mappings with invariant output are commonly used to capture physics invariance principles, where the dependent quantity remains unchanged under certain circumstances. In the following step, we deep dive into how the local search space is defined to keep the transformed test inputs in the neighbourhood of the original data point and constantly meet the foreseeable data conditions. 
\subsection{Inference of Physics-Guided Adversarial Tests}
In this DL adversarial testing, we broaden the scope of adversarial tests to cover not only invariance properties such as imperceptible image perturbations (largely applied to computer-vision models), but also to assert the model's input-output sensitivities with physics-grounded expectations. Given the specified sensitivity rules, we can infer two main types of adversarial tests depending on the expected deviation of the predicted quantity under the conditions on input features: 

\textbf{Invariance Test: }It restricts the input generation to the datapoint-wise local neighbours that should share equal predicted quantity, in order to verify the consistency of the model with experts' invariant-output mappings, reflecting physics invariance principles. The test assertion consists of the equality test between the model's predictions on the original and its derived synthetic data points.

\textbf{Directional Expectation Test: }It applies the constrained perturbation of input features, as outlined in the premises of the specified sensitivity rules. The resulting synthetic neighbours data should lead to the expected directional deviation on the model's predictions, as foreknown by the experts in the conclusion of the sensitivity rules. Thus, the test assertion depends on inferring the outputs of the model for both original datapoint and its craftly-perturbed neighbors, then, makes sure the directional expected deviations happen (i.e., the synthetic datapoints' predictions should be either higher or lower than the original datapoint's prediction).\\
Although the defined adversarial tests rely on sensitivity rules grounded by theoretical physics laws and processes, they would be applied on experimental sensors data that might encouter different sources of noise. Hence, we decide to soften the equations and inequalities involved in the above-mentioned tests assertions by including a tolerable deviation error, $tol_i$, on any measured sensor input, $x_i$, according to domain expert guidances. 
\subsection{Search-based Approach for Physics-Guided Adversarial Testing}
The backbone of the proposed physics-guided adversarial testing is the metaheuristic-based optimization that searches for the erroneous behaviors of the DNN against the specified physics-grounded sensitivity rules. Below, we explain the development steps that we follow to construct this search-based testing approach. 
\subsubsection{Definition of the data search space}
A physics-guided adversarial test may be one of two types depending on the assertion. It is either to assert if the output is invariant, or if the expected directional deviation occurs. The physics-based sensitivity rules specify the assertion type in their conclusions. As a result, the input search space can be defined similarly for both types of physics-guided adversarial tests. From the premises of the underlying sensitivity rule, we derive the space boundaries, which include the upper and lower bounds of each involved input feature. For instance, a sensitivity rule, stating that $x_1\nearrow$ and $x_2\searrow$ $\longrightarrow$ $y\nearrow$, corresponds to the subspace of all the inputs, $\hat{x}$,  where $\hat{x}_1 > x_1$ and $\hat{x}_2 < x_2$. For each inequality in the feature space, however, we still do not have the other side boundary. We could complete the missing boundaries in the above-mentioned inequalities of feature space by using the full range of expectations, $[m_i, M_i]$ for each input feature $x_i$. Hence, the search space of inputs in the example above is $\hat{x}$, where $x_1 < \hat{x}_1 < M_1$ and $m_2 < \hat{x}_2 < x_2$. Other input features not mentioned in the rule's premises are denoted, $x_c$, and they are supposed to remain constant. In order to account for the experimental noise, we set their values range to be between $[min(m_i, x_i – tol_c), max(M_i, x_i + tol_c)]$. The search of the derived input space corresponds to sampling the neighbors of the original datapoints, for which the specified sensitivity rule should permanently apply. As a result, these sampled neighbors datapoints would serve as physics-based adversarial test inputs.
\subsubsection{Definition of the foreseeable data constraints}
Any part of an aircraft must meet its intended performance requirements in all foreseeable operating conditions in order to be certified. United States Code of Federal Regulations (CFR), part $25$, for Jet aircraft design states that:
\begin{displayquote}
\textbf{‘‘the equipment, systems, and installations must be designed and installed to ensure they perform their intended functions under all foreseeable operating conditions.''} 
\end{displayquote}
To simulate aircraft components under different operating conditions of interest, the designed performance models should include input features that reflect flight conditions (e.g., altitude, speed, and outside temperature), in addition to the system-related sensor data. As a result, our test cases should pertain to aircraft flight envelopes and foreseeable ambient conditions in order to represent meaningful and useful simulation scenarios. There are physics processes that govern the interactions between ambient conditions measurements and aircraft operating envelopes, which specify, as an example, the maximum airspeed the aircraft can reach at a given pressure altitude. Therefore, we define data constraints limiting the input search space to the foreseeable ambient and flight conditions. In order to be considered as valid test cases, the sampled inputs must satisfy the defined foreseeable data constraints. In our performance models of aircraft systems, we derived the foreseeable pairs of input features (altitude, airspeed) and (altitude, ambient temperature) from the flight envelopes. Additionally, the operating airspeed is further delimited when high-lift devices are extended (e.g., slats and flaps), according to the aircraft operating conditions. To increase the likelihood of generating inputs that satisfy all of the foreseeable data constraints, the latter are also incorporated into the input generation problem.
\subsubsection{Design of the fitness function}
The fitness function, $fitness(\hat{x})$, should directly measure how much the model’s prediction for the input $\hat{x}$ is inconsistent with the specified physics sensitivity rule, $r_j$. Thus, the generated input data with higher fitness values would have high chances to fail the underlying adversarial test. Given the model $f$ under test, an original input $x$ and its derived synthetic $\hat{x}$ resulting from applying the rule, $r_j$, the deviation from the desired behavior would depend on the type of the adversarial test. Regarding the invariance test, we verify constant-output rules $r_j\in R_{cons}$ where the absolute difference between original and synthetic predictions, $|f(x) - f(\hat{x})|$, can be the behavioral deviation. Concerning the directional expectation test, we validate either increasing-output rules $r_j\in R_{incr}$ or decreasing-output rules $r_j\in R_{decr}$, where the behavioral deviation would be, respectively, $f(x) - f(\hat{x})$ and $f(\hat{x}) - f(x)$. Hence, the fitness function should be equal to the behavioral deviation measure, $fitness(\hat{x}) = dev_{f,r_j}(x,\hat{x})$ that can be formulated as follows:
\begin{equation}
\label{dev_eqs}
dev_{f,r_j}(x,\hat{x}) = \begin{cases} f(x) - f(\hat{x}), & \mbox{if } r_j\in R_{incr} \\ |f(x) - f(\hat{x})|, & \mbox{if } r_j\in R_{cons} \\ f(\hat{x}) - f(x), & \mbox{if } r_j\in R_{decr} \end{cases}
\end{equation}
Using the behavioral deviation measure $dev_{f,r_j}(x,\hat{x})$, we can generalize the assertion of both adversarial test types to be the inequality, $dev_{f,r_j}(x,\hat{x}) > tol_{j}$, where $tol_{j}$ is the tolerance predefined for the output quantity.
\subsubsection{Implementation of metaheuristic-based optimizers}
As introduced in Section \ref{SBST}, we apply SBST, using population-based metaheuristics to drive optimally the data generation towards diverse prominent regions in the input space of the underlying adversarial test. In line with the No Free Lunch Theorem (NFL)~\cite{ho2002simple}, we implement two concurrent population-based metaheuristics, PSO and GA, that are described in \ref{SBST}. Then, we tune each metaheuristic algorithm's hyperparameters to appropriately tune its level of nondeterminism and balance between the intensification (exploiting the results and concentrating the search on regions near effective solutions found) and diversification (exploring non-visited regions to avoid missing interesting potential solutions)~\cite{joshi2020parameter}. In our test data search, we aim for the optimal balance between new test inputs that are sufficiently different from the old ones to uncover new regions of interest while at the same time similar enough to test inputs that have high fitness values to uncover more adversarial examples, which belong to the interesting regions that were previously found. Besides, the enhancement of the fitness values, over iterations, produces new test inputs with higher deviations with respect to the expected behavior, and hence, these test inputs would have likely better chances to expose DNN’s physics inconsistencies. There is, however, no optimal or suboptimal solution that would represent, in our case, the test input with the highest fitness. Consequently, we modify slightly the standard design of population-based metaheuristics to continuously monitor adversarial examples among evolving feasible solutions. A feasible solution is any valid test input that successfully meets all the sets of foreseeable constraints. A valid test input that violates the underlying physics sensitivity rule constitutes an adversarial example.

By completing all the above construction phases, we arrive at the proposed physics-guided adversarial testing workflow shown in Figure~\ref{fig:overview_phys_adv_learning}, which lays out the steps in the following order: \textit{(1)} For each original input, $x^{(i)}$, and sensitivity rule $r_j$, the search space is instantiated according to the premises of $r_j$ and the features’ values of $x^{(i)}$; \textit{(2)} With the current sensitivity rule $r_j$ and the model under test $f$, the deviation function, $dev_{f,r_j}$, would be as in Eq.~\ref{dev_eqs}; \textit{(3)} The population-based metaheuristic algorithm would search over the input space for the most-fitted entries, i.e., the ones triggering high deviation values and satisfying all the foreseeable constraints; \textit{(4)} For all the generated synthetic inputs $\hat{X}_{i,j}$, a follow-up adversarial test would assert if the computed deviation, $dev_{f,r_j}(x_i,\hat{x}_{i,j})$, exceeded a prefixed threshold, which by default equals the rule tolerance. $tol_j$. Each revealed adversarial example, $\hat{x}_{adv}$, should also be stored along with its metadata (parent index $i$, expected deviation $d$, permissible tolerance $tol_j$).\\ 
As the leveraged metaheuristic algorithms are iterative, our workflow encapsulates a nested loop of the steps, \textit{(3)} - \textit{(4)}, that would be repeated for a total of maximum iterations, $K$. Indeed, the searcher starts at the first iteration, $k=0$, with a population of candidate inputs, $\hat{X}_{i,j}^{(k)}$, randomly sampled from the neighbors of $x^{(i)}$ w.r.t the premises of $r_j$. Then, it computes their fitness scores, as follows, \textit{fitness($\hat{x}$) = }$dev_{f,r_j}(x^{(i)},\hat{x})$ $\forall \hat{x}\in\hat{X}_{i,j}^{(k)}$, and captures all the $\hat{x}_{adv}$ that meet the condition of \textit{fitness}($\hat{x}_{adv}$)$>tol_j$. Afterwards, the metaheuristic update routines evolve the population, $\hat{X}_{i,j}^{(k)}$, and derive new candidates, $\hat{X}_{i,j}^{(k+1)}$, that are probably better than their predecessors in terms of fitness. Thus, the main loop of our workflow consists of running all of the steps (\textit{1}$\rightarrow$ \textit{4}) for all the original inputs, $X$, as well as for all the applicable sensitivity rules $R$. Thus, the semi-supervised adversarial examples, $\hat{X}_{adv}$, would serve us later in fine-tuning the DNN's mappings to capture the underlying physics processes and system-related properties.

\section{Physics-informed Adversarial Training}
\label{sec:adv_training}
Regularization penalties are often developed by ML scientists that reflect model complexity and encourage optimizers to find simpler mapping between features and outputs. Thus, the loss function would be written as follows, $l(y, \hat{y}) + \lambda . R(\theta)$. The widely-used regularization penalties are $L_1$-norm and $L_2$-norm on the parameters, which respectively impose low-magnitude parameters and sparse parameters (i.e., with zeroed coefficients). In this section, we propose a physics-informed regularization by devising a penalty that encourages the training algorithm to maintain a reasonable level of physics consistency in the learned mapping function. 
\subsection{Physics-informed Regularization Cost}
A follow-up fine-tuning procedure will leverage revealed adversarial examples to fix these erroneous behaviors, similarly to conventional adversarial learning~\ref{adv_train}. Nevertheless, it is natural to retrain using the semantically-preserving adversarial examples as we expect no change in class label or slight deviation from continuous output. In contrast, we need to be prepared for a deviation in the predicted outputs in response to our produced synthetic inputs, $\hat{x}$, but we are aware of the correct direction of the change in the model's predictions, based on the physics-based sensitivity rule, $r_j$. As a result, we introduce a variable $d$ equals to $1$ or $-1$ if we expect the output to increase ($r_j\in R_{incr}$) or decrease ($r_j\in R_{decr}$), respectively, and $d=0$ if we do not expect any significant change in the output ($r_j\in R_{cons}$). Therefore, we aim to design a physics-informed regularization that uses the semi-supervised, generated test data assembling the transformed inputs $\hat{x}$, the expected change direction $d$, and the reference pointed to its parent input $x$. As a first step, we were inspired by the \textit{hinge\_loss}$=(0, 1 -f(x).y)$, which is commonly used for maximum-margin classifiers~\cite{hinge_loss}, such as support vector machines (SVMs). The hinge loss is a specific type of cost function that incorporates a margin from the hyperplane, representing the classification decision boundary. Thus, it penalizes even correctly-classified data points if they are very close to the hyperplane, i.e., their distances are less than the margin. In addition to placing the data points on the correct side of the hyperplane, the hinge loss encourages the classifier to place them beyond the margin as well. Indeed, the distance from the hyperplane can be regarded as a measure of confidence. Therefore, we estimate the physics-consistency error to reflect how far the model deviates from the expected change direction, using the deviation function (Eq.~\ref{dev_eqs}). Then, we ignore any noise-induced deviations within the margin of error by using the domain-specific tolerance, $tol$, defined by experts. Afterwards, we square the non-zero deviations to give a stronger weighting to larger differences and to follow the same scale as the squared prediction errors. The resulting regularization cost for the model $f$ under test can be formulated in the below Eq.~\ref{R_eqs}.
\begin{small}
\begin{equation}
\label{R_eqs}
R_{phys}(x, x_{adv}) = \begin{cases} (max(tol, f(x) - f(\hat{x})) - tol)^2, & \mbox{if } d = 1 \\ max(tol^2,(f(x) - f(\hat{x}))^2)-tol^2, & \mbox{if } d = 0 \\ (max(tol,f(\hat{x}) - f(x)) - tol)^2, & \mbox{if } d = -1 \end{cases}
\end{equation}
\end{small}
where $x_{adv}=(\hat{x}, d, tol)$, $x, \hat{x} \in \mathbb{R}^D$, $d\in\{-1,0,1\}$, $tol\in \mathbb{R}_+$.\\
Then, we generalize the regularization cost in a single generic function that includes all the types of our physics-based adversarial test assertions, as below formulated in Eq.~\ref{R_general}.
\begin{small}
\begin{equation}
\label{R_general}
R_{phys}(x, x_{adv}) = [max(tol^{p_2}, ((-1)^{p_3}(f(x)-f(\hat{x})))^{p_2}) - tol^{p_2}]^{p_1} 
\end{equation}
\end{small}
where $p_1=1+d^2$, $p_2=2-d^2$, $p_3=|d|(\frac{d+1}{2}+1)$.\\
Furthermore, we would emphasize the importance of including a regularization tolerance, $tol$ as a predefined hyperparameter, which can be equal, by default, to the tolerance used in adversarial detection to overcome experimental noises. In the absence of this tolerance, the physics-informed regularization would likely push the model towards having, empirically, zero cost in terms of violating theoretical physics rules, but based on experimental data. Hence, we run the risk of stagnation and perhaps even divergence as we attempt to strictly apply the physics rules to noisy data.
\subsection{Physics-informed Adversarial Training Algorithm}
For the sake of simplicity, we have presented our physics-informed regularization cost estimation on a pair of original, $x$, and adversarial inputs, $\hat{x}$. To avoid sub-optimal local minima, we usually use mini-batch stochastic optimization for training DNNs. So, we will also apply a mean reduction strategy to the regularization costs of all inputs in a batch. The mean over the sum reduction is chosen because it preserves the invariance of the loss scale to the batch size, and it is aligned well with the mean squared error, commonly used as data loss for regression models. Therefore, the physics-informed loss function that sums the original loss and the cost of its integrated regularization for batches of data ($X^b$, $y^b$) can be defined as follows:
\begin{equation*}
l_{phys}(X^b, y^b, X_{adv}^b) = l(f(X^b),y^b) + \lambda_{phys} R_{phys}(X^b, X_{adv}^b)
\end{equation*}
where $X_{adv}^b = (\hat{X}^b, d, tol)$, $x\in \mathbb{R}^{B\times d}$, $\hat{X}^b \in \mathbb{R}^{M\times d}$, $d\in\mathbb{R}^M$, $tol\in \mathbb{R}_+^M$, $\lambda_{phys} \in \mathbb{R}_+$ controls the strength of the physics-informed regularization penalty.\\
In our preliminary experiments, we observed that the choice of lambda, $\lambda$, could be challenging and hinder the performance of physics-informed adversarial training. Below we describe the two identified major challenges in lambda, $\lambda$, setup with respect to the substantial differences in magnitude of both costs (namely, data loss and regularization).\\
\textbf{Unfair cold start conditions between both costs: }In fine-tuning, we start with pre-trained DNN and its corresponding adversarial examples that highlight its revealed inconsistencies with the underlying physics sensitivity rules. As a consequence, it is natural to have initially low data loss, $l(f(X^{(b)}),y^{(b)})$, and a high physics-informed regularization cost, $R_{phys}(X^{(b)}, X_{adv}^{(b)})$. Thus, the fine-tuning process focuses primarily on minimizing physics-informed regularization cost, while being excessively tolerant of substantial increases in data loss, as long as the total sum-up loss, including the regularization penalty, continues to decrease. As a solution to this starting bias, we set lambda, $\lambda$, in a way that aligns the magnitudes of both costs from the beginning (i.e., the first iteration) in order to start from fair initial conditions.\\
\textbf{Substantially different sizes of adversarial batches over iterations: }Despite the use of mean reduction in each cost to normalize over the batch data, we face cases where the original batch has no corresponding adversarial examples or may have several of them. This induces substantial differences between the batch data loss, $l(f(X^b),y^b)$, and its related physics-informed penalty cost, $R_{phys}(X^b, X_{adv}^b)$; so we dynamically update the lambda value, $\lambda_{phys}$, in order to adapt the magnitude of the additional penalty cost depending on its initial order of magnitude, whenever it exists (i.e., not null). As a reference loss value, we use the average loss estimated over all the batches using the best fitted model before starting the fine-tuning, because the batch losses could substantially differ too and destabilize the convergence.\\
To illustrate how the physics-informed adversarial training algorithm works, we describe the remaining steps of the workflow in Figure~\ref{fig:overview_phys_adv_learning} and we refer to the lines of code from the Algorithm’s Pseudo Code~\ref{pseudo_code}: \textit{(5)} The algorithm guarantees non-divergence from the best-fitted state as it iterates over batches of original data, $X^{(b)}$, (code lines 6-7) and watches the model’s loss on them (code line 13), which means there is no substantial degradation in regards to the fit of the original distribution; \textit{(6)} It searches all AXs, $\hat{X}_{adv}^{(b)}$, and their metadata including expected change directions $D^{(b)}$ and deviation tolerance $T^{(b)}$, which have been revealed relying on the original input, $X^{(b)}$ (code line 10), and computes their corresponding physics-informed regularization cost, $R_{phys}(X^b, X_{adv}^b)$ (code line 15); \textit{(7)} It dynamically calibrates the magnitudes of both losses (code line 17), by inferring the lambda coefficient, $\lambda$, (code lines 22-26) to keep the regularization cost aligned with the base loss (i.e., the model's average data loss at launch).
\begin{lstlisting}[caption={Physics-Based Adversarial Training Algorithm},label=pseudo_code, language=Python]
#prepare the base data loss as the model loss before fine-tuning
preds = DNN(X)
base_data_loss = MSE(y, preds)
for epoch in epochs:
   indices, X, y = shuffle(X, y)
   batches = data_loader(indices, X, y, batch_size)
   for indices_b, X_b, y_b in batches.iterate():
        preds_b = DNN(X_b)
        #search all the adversarial examples connected to the current batch entries
        X_adv, d_adv, t_adv = adversarial_data.search(indices_b)
        preds_adv = DNN(X_adv)
        #compute the data loss
        data_loss = MSE(y_b, preds_b)
        #compute the physics regularizatio cost
        reg_cost = R_phys(preds_b, preds_adv, d_adv, t_adv)
        #compute dynamically lamda to align the magnitudes of both losses
        lamda = pow(10, magnitude_order(base_data_loss/reg_cost))
        #aggregate both losses
        loss = data_loss + lamda * data_loss
	    DNN = update_model(DNN, loss)
	    
def magnitude_order(value):
    '''computes the magnitude order of a real value.
       For example, (0.004) --> (-3), (105) --> (2)
    '''
    return math.floor(math.log(value, 10))
\end{lstlisting}
\section{Evaluation}
\label{sec:evaluation}
In this section, we introduce two industrial case studies, as well as our evaluation setup, metrics, and methodology. Next, we test the effectiveness of our physics-guided adversarial machine learning approach for assessing and improving the trustworthiness of neural networks used to simulate aircraft performance.
\subsection{Experimental Setup}
\subsubsection{Case Studies}
Two aircraft systems performance simulation models were used in our empirical evaluation. The first case study consists of an aircraft performance (referred to as \textit{A/C Perf.}) model mapping steady-state angle of attack ($\alpha$) to features related to flight conditions and wing configurations. The second case study consists of an in-flight wing anti-icing performance (known as \textit{WAI Perf.}) model mapping wing leading-edge skin temperature sensors to features related to flight conditions, wing configurations, and high-pressure pneumatic system conditions at the wing root. Indeed, Jet aircraft often use hot-air ice protection systems that prevent ice accumulation over the wings during flight. Thus, the performance of the In-flight wing anti-icing system is determined by its ability to sustain, under all foreseeable conditions, a wing leading-edge skin temperature sufficient to prevent ice formation. These case studies were run on the regression data sets summarized in the Table~\ref{tab:size-dim-dataset} where $N$ is the number of steady state flights and $D$ is the number of dimensions.
\begin{table}[h]
\centering
\caption{Size and dimensionality of data sets}
\label{tab:size-dim-dataset}
\begin{tabular}{|l|l|l|}
\hline
\textbf{Dataset} & \textbf{N} & \textbf{D}  \\ \hline
Aircraft Performance Data   & 1338 & 5  \\ \hline
Wing Anti-Icing Performance Data & 90   & 10 \\ \hline
\end{tabular}
\end{table}
The problem-related data (features and targets) as illustrated in Table~\ref{tab:AOA-reg-data-catalog} and~\ref{tab:WAIS-temp-reg-data-catalog}, were obtained from flight test campaigns and were provided as a dataframe, with entries representing distinct steady-state flights, and columns representing selected features. 

\begin{table}[h]
\centering
\caption{Aircraft performance Data Catalog}
\label{tab:AOA-reg-data-catalog}
\begin{tabular}{l|l|l|}
\cline{2-3}
& \textbf{Name}  & \textbf{Type}  \\ 
\hline
\multicolumn{1}{|l|}{
\multirow{4}{*}{\textbf{Features}}} & Calibrated Airspeed& Real\\ 
\cline{2-3} 
\multicolumn{1}{|l|}{} & Aircraft Weight& Real\\ 
\cline{2-3} 
\multicolumn{1}{|l|}{} & Slat Angle& Dichotomous\\ 
\cline{2-3} 
\multicolumn{1}{|l|}{} & Flap Angle& Categorial\\
\hline
\multicolumn{1}{|l|}{\textbf{Target}} & Angle of Attack& Real\\ 
\hline
\end{tabular}
\end{table}

\begin{table}[h]
\centering
\caption{WAI Performance Data Catalog}
\label{tab:WAIS-temp-reg-data-catalog}
\begin{tabular}{l|l|l|}
\cline{2-3}
& \textbf{Name}& \textbf{Type} \\ \hline
\multicolumn{1}{|l|}{\multirow{8}{*}{\textbf{Features}}} & Angle of Attack & Real\\ \cline{2-3} 
\multicolumn{1}{|l|}{}                                   & True Airspeed& Real\\ \cline{2-3} 
\multicolumn{1}{|l|}{}                                   & Pressure Altitude& Real\\ \cline{2-3} 
\multicolumn{1}{|l|}{}                                   & Pressure of Wing Root& Real\\ \cline{2-3} 
\multicolumn{1}{|l|}{}                                   & Temperature of Wing Root& Real\\ \cline{2-3} 
\multicolumn{1}{|l|}{}                                   & Total Air Temperature& Real\\ \cline{2-3} 
\multicolumn{1}{|l|}{}                                   & Slat Angle & Dichotomous \\ \cline{2-3} 
\multicolumn{1}{|l|}{}                                   & Flap Angle& Categorical\\ \hline
\multicolumn{1}{|l|}{\multirow{2}{*}{\textbf{Target}}} & 
A-Wing's Skin Temperature Sensor & Real \\ 
\cline{2-3} 
\multicolumn{1}{|l|}{}&  B-Wing's Skin Temperature Sensor & Real \\ 
\hline
\end{tabular}
\end{table}

Indeed, these steady-state flight points are derived and preprocessed meticulously from costly flight test time series. They represent various flight states in which conditions must remain stable over time for our physical assumptions to hold (i.e. steady-state conditions). As can be observed, the number of flight datapoints may differ substantially due to the difficulty of relevant steady state extraction. In our study cases, the A/C performance model lies on standard flight parameters (i.e., speed, altitude, and weight), which are often found to be stable during flight. Hence, we were able to collect numerous datapoints that correspond to conditions commonly encountered in flight test campaigns. On the other hand, the steady-state datapoints for WAIS performance modeling are quite difficult to extract. They require the aircraft pilot to fly for quite a long time (i.e, up to tens of minutes) during which the WAI system is precisely configured. Given the high cost of flight tests, aircraft engineers often require such limited-size preprocessed steady-state flight to conduct aircraft subsystems engineering analysis. They employ extensive advanced analytics combined with their domain knowledge to provide the necessary demonstrations for certification with the minimal data points. This is known as critical point analysis (CPA). For example, certification of engine icing protection mechanisms~\cite{cfr} requires the analysis of the most critical points that provide evidence of the systems' effectiveness across the entire ice envelope under all foreseeable operating conditions (hold, descent, approach, climb, and cruise). With either a limited set of critical points or a high coverage of normal operating conditions, it is challenging to design and fit a data-driven simulation model that can predict system behaviors under all the foreseeable conditions in the flight envelope. 
To circumvent the limited size of datasets, the engineering team defined system-centric transformations, as outlined in Table IV. Given the limited range of variation $t$ within [-10, 10], the one-to-many augmentation rule, $A_1$, mimics different equilibrium states of temperatures based on the genuine flights to enhance the diversity among the training samples. The one-to-one augmentation rules, $A_{2}$, $A_{3}$ and $A_{4}$, serve as boundary conditions that help inject artificial data points with the aim of improving the model's data fitness. In the following, we describe the augmentation rules from Table~\ref{tab:Aug_WAIS_rules} and their rationales.
\begin{table}[h]
\renewcommand{\arraystretch}{1.5}
\centering
\caption{Augmentation Rules for WAIS Perf. Model}
\label{tab:Aug_WAIS_rules}
\begin{tabular}{|l|l|l|l|l|}
\hline
\textbf{Status} & \textbf{Rule} & \textbf{Premises} & \textbf{Conclusion} \\ \hline
\multirow{3}{*}{ON}&\multirow{2}{*}{$A_1$} & TAT $+$ $t$ , TWR $+$ $t$ , & 
\multirow{2}{*}{$T_{skin}^{a,b}$ $+$ $t$} \\ 
&&$\forall t\in[-10, 10]$ & \\ \cline{2-4} 
&$A_2$ & PWR $= 0$ & $T_{skin}^{a,b}$ $=$ TAT \\ \cline{2-4} 
&$A_3$ & TAS $= 0$ & $T_{skin}^{a,b}$ $=$ TWR \\ \hline
OFF& $A_3$ & TWR $=$ TAT & $T_{skin}^{a,b}$ $=$ TAT\\ \hline
\end{tabular}
\end{table}
The expected variation of leading-edge skin temperatures, $T_{skin}^i$, $i\in{a,b}$, must be approximately equal to the variations of temperature at wing (TWR) and the total air temperature (TAT) when the later is equal and within a predefined range. In $A_1$, aircraft engineers have fixed an absolute difference less than $10$. In $A_2$, the rule sets the PWR to 0, eliminating the flow of WAI through the wing, which consequently cancels the effect of TWR on the skin temperatures. Thus, they become equal to TAT. In $A_3$, the rule simulates synthetic data points at extreme conditions, by attributing 0 to the airspeed (TAS). Theoretically, this results in no forced convection and the wing skin cannot transfer heat to the outside air. Thus, both $T_{skin}^a$ and $T_{skin}^b$ should be equal to TWR. In addition, the augmentation rule, $A_4$ can be applied when WAI is deactivated. It sets TWR to be equal to TAT; so TAT and TWR are set to the same value. Accordingly, the two leading-edge skin temperatures must be approximately equal to this value. For further details on the heating and cooling principles governing the WAI system, we refer to our explanation of its physics-grounded sensitivity w.r.t each input quantity in Section~\ref{phys-sensitivity-section}. Thereby, the training data loader incorporates an online augmentation step that randomly transforms the mini-batches feeded to the on-training model. It helps prevent overfitting as the model rarely encounters the exact same inputs multiple times and cannot simply memorize them. Our online augmentation randomly applies one of the defined augmentations half of the time, i.e., a genuine training input can be transformed with a probability of $0.5$.
As certain simulation-driven risk analyses are conducted under extreme and rare conditions with low coverage, our physics-based adversarial ML leverages both system-related and physics domain knowledge to validate and improve the physics consistency of the statistically-learned models; this makes them more viable alternatives to purely physics-based simulators. In addition, the flight envelopes for the studied Jet aircrafts, which included the parameters of interest, were gathered from the internal specifications of Bombardier aircraft products. 
\subsubsection{Physics-based Sensitivity Rules}
\label{phys-sensitivity-section}
In the following, we briefly expose the physic-based high-level requirements for the above-mentioned systems in terms of input-output sensitivity rules using expert know-how. The specifications of the rules are inferred from a detailed knowledge of the aircraft performance principles~\cite{eshelby2000aircraft} and the local heat transfer characteristics~\cite{hahn2012heat} that arise from energy, momentum and mass balance over an aircraft wing.\\
\textbf{A/C Performance Model. }The performance model of the studied aircraft is associated to the fundamental performance characteristics of an aircraft. It associates the aircraft orientation to the amount of weight it can lift at a certain speed. The model uses calibrated airspeed (CAS), a quantity that is by definition strongly correlated to the true airspeed. The model also assumes non-accelerated flight at small flight path angles; hence the lift is approximately equal to the aircraft weight (ACWT). From the laws of motion and basic knowledge of air flow around thin bodies, one can infer several sensitivity relations between system’s inputs and output. First, one can state from Newton’s third law that aircraft wings create an upwards force by deflecting air flow downwards. When airspeed increases, less deflection is necessary to produce the same force. This relation holds true for thin, smooth profiles at low deflection angles, for which the airflow deflection is assumed to be strongly correlated to the angle of attack ($\alpha$). Thus, $\alpha$ is inversely correlated to CAS at constant ACWT. Second, Newton’s third law also justifies the assumption that greater air flow deflection generates more lift, allowing to carry more weight. Therefore, $\alpha$ is positively correlated to ACWT at constant CAS. Third, aircraft’s slats and flaps are devices designed to increase lift when flying at low speed. Hence, it is expected to observe operationally a lower aircraft $\alpha$ as these high-lift devices are deployed. Therefore, in normal operations, $\alpha$ is expected to be negatively correlated to the slat deployment bit, and to the flap angle. These three a priori observations are then encoded into physics-based sensitivity rules, as shown in Table~\ref{tab:AOA_rules}.\\
\begin{table}[h]
\centering
\caption{Physics-grounded Sensitivity Rules for A/C Perf. Model}
\label{tab:AOA_rules}
\begin{tabular}{|l|l|l|}
\hline
\textbf{Rule}  & \textbf{Premises}    & \textbf{Conclusion}  \\ \hline
$r_0\in R_{incr}$ & CAS $\downarrow$ , ACWT $\uparrow$ , Flap $\downarrow$ , Slat $\downarrow$ & $\alpha$ $\uparrow$ \\ \hline
$r_1\in R_{decr}$ & CAS $\uparrow$, ACWT $\downarrow$, Flap $\uparrow$, Slat$\uparrow$ & $\alpha\downarrow$ \\ \hline
$r_2\in R_{cons}$ & CAS $\leftrightarrow$ , ACWT $\leftrightarrow$ & $\alpha$ $\leftrightarrow$ \\ \hline
\end{tabular}
\end{table}
\textbf{WAI Performance Model. }During flight, the anti-icing capability of jet aircraft wings is determined by the control of the leading-edge skin temperature, measured as $T_{skin}^i$, $i\in{a,b}$, which prevents ice from forming on the wings. For hot-air anti-icing systems, adequate skin temperatures are assured by the hot air flow stream entering the wing at temperature (TWR) that compensate for the loss of temperature caused by the ambient air surrounding the wing (at temperature TAT). As a matter of fact, an increase of TAT and TWR will result in warmer skin temperatures. Based on energy conservation and convection heat transfer~\cite{incropera1996fundamentals}, the efficiency of the energy exchange between the internal and external air streams can be encoded with the following sensitivity rules with focus on some key variables. First, an increase of WAI pressure at the wing root (PWR) induces increased internal air flow to raise the skin temperatures, $T_{skin}^a$ and $T_{skin}^b$. Then, a higher aircraft air speed (TAS) eases the heat exchange with the outside air, which pushes skin temperatures closer to those of the outside air stream. Inversely, an elevated altitude (ALT) causes the air density to decrease along with the heat exchange with the outside, which consequently raises both of $T_{skin}^a$ and $T_{skin}^b$. Also, these skin temperatures are negatively correlated to the angle of attack ($\alpha$) which affects the pressure and airflow temperature distributions above the wings, causing fluid to accelerate more rapidly in the upper side wing areas. When the WAIS is turned off, the effects of input features, namely, TAS, ALT, and TWR become negligible, which results in more compact sensitivity rules with less variables in the premises. A wing skin temperature, $T_{skin}^i$, $i\in{a,b}$, reaches the total air temperature (TAT) at steady-state if WAIS is deactivated. Hence, varying the aircraft's speed or altitude does not directly affect the skin temperatures, $T_{skin}^a$ and $T_{skin}^b$, but they influence the atmospheric conditions outside the aircraft. We already account for these effects with the concept of total air temperature (TAT). Furthermore, they are not affected by variations in TWR, since there is no flow of WAI passing through the wing. Last, the slat extension (SLAT) and flap angle (FLAP) were discarded in the sensitivity rules for the wing anti-icing system (WAIS) because their effects on skin temperatures are difficult to characterize. SLAT and FLAP variations induce local changes in the flow behavior over the wing surface, but the resulting sensitivity of $T_{skin}^i$, $i\in{a,b}$ depends on the exact location of the wing temperature sensor. Characterizing the sensitivity would require advanced modeling based on a precise analysis of the local flow patterns with computational fluid mechanics modeling methods. Therefore, Table~\ref{tab:WAIS_rules} summarizes all the physics-based sensitivity rules that can be exploited by our approach on WAI performance model's during a physics-guided test session.\\ 
\begin{table}[h]
\renewcommand{\arraystretch}{1.5}
\centering
\caption{Physics-grounded Sensitivity Rules for WAIS Perf. Model}
\label{tab:WAIS_rules}
\begin{tabular}{|l|l|l|l|l|}
\hline
\textbf{Status} & \textbf{Rule} & \textbf{Premises} & \textbf{Conclusion} \\ \hline
\multirow{2}{*}{ON}      & \multirow{2}{*}{$r_3\in R_{incr}$ } & ALT $\uparrow$ , TWR $\uparrow$ , PWR $\uparrow$ , & \multirow{2}{*}{$T_{skin}^{a,b}$ $\uparrow$} \\ 
&&TAT $\uparrow$ , TAS $\downarrow$ , $\alpha$ $\downarrow$ &  \\ \cline{2-4} 
                             & \multirow{2}{*}{$r_4\in R_{decr}$} & ALT $\downarrow$ , TWR $\downarrow$ , PWR $\downarrow$ , & \multirow{2}{*}{$T_{skin}^{a,b}$ $\downarrow$} \\ 
                          && TAT $\downarrow$ , TAS $\uparrow$ , $\alpha$ $\uparrow$ & \\ \hline
\multirow{2}{*}{OFF}   & $r_5\in R_{incr}$  & TAT $\uparrow$ , PWR $\uparrow$ , $\alpha$ $\downarrow$  & $T_{skin}^{a,b}$ $\uparrow$  \\ \cline{2-4} 
                           & $r_6\in R_{decr}$ &  TAT $\downarrow$ , PWR $\downarrow$ , $\alpha$ $\uparrow$  & $T_{skin}^{a,b}$ $\downarrow$     \\ \hline
\multirow{2}{*}{Both} & \multirow{2}{*}{$r_7\in R_{cons}$} &  ALT $\leftrightarrow$ , TWR $\leftrightarrow$ , PWR $\leftrightarrow$ , & \multirow{2}{*}{$T_{skin}^{a,b}$ $\leftrightarrow$} \\
&& TAT $\leftrightarrow$ , TAS $\leftrightarrow$ , $\alpha$ $\leftrightarrow$&\\ \hline
\end{tabular}
\end{table}
Both system models lack physics-grounded invariant rules, but we added a constant-output sensitivity rule for each model, $r_2$ and $r_7$, that describes the degree to which the model's output can change as input features change. Following expert guidance, domain-specific tolerances are used to define the radius of perturbations on the inputs and the maximum unsigned deviations of the outputs. Indeed, the invariance tests are essential to ensure the smoothness of the learned mapping function as well as its numerical stability to be a viable simulation solution for one of these smooth or piecewise-smooth dynamical systems.
\subsubsection{Models}
Our base nonlinear regression model is a feedforward neural network(FNN) that is trained using the Mean Squared Error (MSE) loss function with L2-norm regularization. Rectified linear units (ReLU) are used as hidden layer activation functions, and Adam is used as the optimization algorithm. In regards to the architecture, we followed the design principle of pyramidal neural structure~\cite{design_principles}, i.e., from low-dimensional to high-dimensional feature spaces/layers, as well as these dimensions are powers of $2$ to achieve better performance on GPUs~\cite{nvidia}. Regarding the preprocessing, both of data inputs and outputs are scaled to have zero mean and unit standard deviation before getting used by the training algorithm in order to standardize their units.\\ 
\textbf{A/C Performance FNN. }It assembles three layers with, respectively, $128$, $64$, and $32$ neurons. The best-fitted parameters were obtained by a training of $250$-epochs using a batch size of $64$, a learning rate of $1e-4$ and an L2-norm weight decay coefficient of $5e-4$.\\
\textbf{Wing-anti Icing Performance FNN. }It stacks the consecutive layers including $256$, $128$, and $64$ neurons. The best-fitted state was reached with $300$ epochs of training using a batch size of $16$, a learning rate of $1e-3$ and an L2-norm weight decay  coefficient of $1e-4$.\\ 
Next, we describe the hyperparameters tuning strategy we adopt to find the above FNNs for our case studies.
\subsubsection{Hyperparameters}
For the FNNs hyperparameters tuning, we leverage the random search strategy to sample several trials of the settings, and we use out-of-sample bootstrap validation that enables stable estimations for relatively small datasets, in order to infer the expected predictive performance of each configuration. A model is trained using a bootstrap sample (i.e., a sample that is randomly drawn with replacement from a dataset) and tested using the rows that do not appear in the bootstrap sample. Indeed, we use a $100$-repeated out-of-sample bootstrap process, where the resampling with replacement is repeated $100$ times. To outline the ranges of the different tuned hyperparameters, we denote $linspace(a, b, n)$ to indicate the range of $n$ equi-spaced values within $[a, b]$ and $logspace(c, d, base)$ to indicate the interval of ${base^c, base^c+1, .., base^d}$, where $c < d$. Starting with the capacity-related hyperparameters, the depth of the neural network is selected from $linspace(1, 6, 1)$, and the size of layers is sampled from the binary logarithmic space, formulated as $logspace(5, 10, 2)$, which keep the enabled learning capacity for the FNNs in accordance with the actual complexity of our case studies. Then, the remaining hyperparameters were tuned as follows: learning rate $\eta \in s \cup 3\times s$, weight decay $\lambda \in s \cup 5\times s$, where $s = logspace(1, 5, 10)$. Batch size was tuned in $logspace(3, 7, 2)$, and epochs count in $linspace(50, 500, 50)$. Concerning the hyperparameters that control the exploration-exploitation trade-off of the involved nature-inspired population-based metaheuristics, we opt for grid search strategy to assess their performances under different alternative settings and find the best configuration in terms of the total count of the revealed unique valid inputs. For PSO, $w$, $\varphi_p$, and $varphi_g$ were tuned in $linspace(0.1, 0.95, 0.05)$, which lead us to set up PSO-enabled test generator with $w=0.8$, $\varphi_p=0.65$, and $varphi_g=0.75$. For GA,  $p_{mutation}$ and $r_{parents}$ were tuned, respectively, in $linspace(0.1, 0.95, 0.05)$ and $linspace(0.1, 0.5, 0.05)$, which ends up with the configuration of $p_{mutation}=0.25$ and $r_{parents}=0.3$. In addition, several binary crossovers for breeding were tested such as one-point~\cite{picek2010comparison}, two-point~\cite{picek2010comparison}, or uniform~\cite{picek2010comparison}, and the one-point crossover operation outperforms the others for our test input generation problem. Last but not least, we rely on system engineering experts' judgment and the operating specifications of the aircrafts used in the flight test data in order to fix the tolerances and the confidence intervals in relation with the features and targets in our case studies.
\subsubsection{Evaluation Strategy and Metrics}
Due to the limited-size of flight test data for both studied system performance, we adopt a $10$-fold cross-validation method for all experiments in order to have different splits for the training and validation datasets and quantify the target metrics by averaging their values over the $10$ iterations. Besides, all the included estimated metrics are computed as average values over $5$ runs or more, in order to mitigate the effects of randomness inherent in metaheuristic search algorithms. Below, we introduce the different evaluation metrics that have been used in the empirical evaluations.\\
\textbf{\%ValIn. }It represents the ratio of valid inputs with respect to the total of all generated inputs. Valid inputs are those that comply with all the foreseeable constraints that encode nonlinear interactions between the input features, as determined by the aircraft operating conditions and atmospheric conditions.\\
\textbf{\%DupIn. }It represents the ratio of duplicate inputs with respect to the total of all adversarial inputs. In our semi-supervised adversarial datasets, duplicate entries are defined by a pairwise Euclidean distance, i.e., two real-valued input vectors with a distance of zero are considered to be duplicates.\\
\textbf{\#AdvIn. }It consists of the number of adversarial inputs, i.e., those that contradict our physics-based sensitivity rules based on their corresponding deviation function, as formulated in Eq.~\ref{dev_eqs}.\\
The following are two metrics based on Percentage Change, which represents the degree of change relative to a base starting point. It can be the percentage of either increase or decrease, which are basically the amount of, respectively, increase or decrease, from the original quantity to the final one in terms of $100$ original parts.\\ 
\textbf{\%Improv\_AdvIn.} The improvement we achieve by regularizing can be estimated using the percentage of decrease in the revealed adversarial test inputs from the original model to its regularized counterpart, as formulated in the below Eq.~\ref{improv_advin}.\\
\begin{equation}
\label{improv_advin}
\%Improv\_AdvIn = \dfrac{\textit{pre-\#AdvIn} - \textit{post-\#AdvIn}}{\textit{pre-\#AdvIn}} \times 100
\end{equation}
Where \textit{pre-\#AdvIn} and \textit{post-\#AdvIn} refer to the actual number inconsistencies (\#AdvIn) revealed before, i.e., using the original version of FNN and after the physics-informed adversarial learning, i.e., the regularized version of FNN.
\textbf{RMSE.} It stands for Root Mean Square Error, which averages all the quadratic deviations between the predicted values and the true/observed ones, and then computes its square root to have an error measured on the same scale as the output. Deviations are proxies for expected prediction errors, since the estimations are based on an out-of-sample, unseen test dataset. In general, RMSE is non-negative and lower values are better than higher ones.\\
\textbf{\%Change\_RMSE.} It consists of the percentage increase, as formulated in the Eq.~\ref{chg_rmse}, because it is unknown in advance how physics-informed regularization will affect prediction errors. Thereby, positive values (highlighted in \textcolor{red}{red}) indicate the on-watch metric, RMSE, increase whereas negative values (shown in \textcolor{dkgreen}{green}) indicate RMSE decrease.\\
\begin{equation}
\label{chg_rmse}
\%Change\_RMSE = \dfrac{\textit{post-RMSE} - \textit{pre-RMSE}}{\textit{pre-RMSE}} \times 100
\end{equation}

To obtain domain experts’ feedback, we recruited two senior engineers from our industrial partner, Bombardier Aerospace: The first engineer works in the aircraft performance team and provides us with feedback regarding the first study case. The second engineer is a specialist in the thermodynamics engineering team and has the expertise required for the second study case. Afterwards, we interviewed them separately to gather their opinions on the strengths/weaknesses of the proposed physics-guided adversarial machine learning, and their experience on the usage of the web-based interface and its provided configurable parameters.
\subsubsection{Software}
The physics-guided adversarial learning framework was developed in Python. It uses Pytorch~\cite{paszke2019pytorch}, an established DL framework for modelling and training neural networks. In addition, our implemented search-based techniques were based on an adaptation of the open-source python libraries, pyswarm\footnote{https://pythonhosted.org/pyswarm/} and geneticalgorithm\footnote{https://pypi.org/project/geneticalgorithm}, to meet the specifications of our designed population-based metaheuristic search algorithms. As a user-friendly version for the collection of domain experts' feedback, we develop a web-based interface for the physics-guided adversarial ML approach in order to facilitate the live user tests. 
\subsection{Experimental Resuls}
We assess the effectiveness of the proposed approach using the following research questions:

\textbf{\textit{RQ1. How effective is the approach at detecting physics inconsistencies?}}

\textbf{Motivation. }The objective is to assess how effective our approach is at testing the physics consistency of a DNN by exposing adversarial examples that highlight problematic regions in the input space, where the model deviates from the foreknown domain knowledge.

\textbf{Method. }We experiment our testing approach using different sizes of data generations, i.e., total of generations $G$ equals to the product of generations count, maximum of iterations, the number of rules, and the number of original inputs, while counting and storing the found adversarial examples. Then, we run the physics-guided adversarial testing on both performance models at each iteration of the cross-validation process, using training and validation dataset splits. Thus, all the estimated counts would be an average of the obtained adversarial examples over the separate cross-validation splits. Besides, we turn off our metaheuristic-based searching algorithm by switching to randomly sampling inputs from the search space of the underlying sensitivity rule, and verifying their validity against the foreseeable data constraints. This input random sampler (RS) represents our simple and inexpensive baseline to assess the added value of the metaheuristic-based searching algorithms in improving the fitness of test inputs over the course of generation. In order to ensure a fair comparison, we sampled a total of random inputs equal to the previously-calculated size of data generations for the studied nature-inspired metaheuristics, PSO and GA.

\textbf{Results.} Table~\ref{tab:RQ1} shows the occurrences counts of unique adversarial inputs (i.e., their predictions are inconsistent with physics-based sensitivity rules) revealed by each search-based method for each performance model, each dataset split, as well as increasing trial sizes.

\begin{table}[ht]
\centering
\caption{The number of unique exposed adversarial cases on train and test datasets per search method and system}
\label{tab:RQ1}
\begin{tabular}{|l|l|l|l|l|}
\hline     
\textbf{SYS}        & \textbf{Method}         & \textbf{G} & \textbf{Dataset} & \textbf{Avg. \#AdvIn} \\ \hline
\multirow{12}{*}{A-C Perf.} & \multirow{4}{*}{RS} & 361K   & $D_{train}$& 1302.3 \\ \cline{4-5}                            &                         & 40K   & $D_{test}$& 128.1 \\ \cline{3-5} 
                       &                         & 903K   & $D_{train}$& 3186.4 \\ \cline{3-5}                            &              & 100K    & $D_{test}$& 340.3 \\ \cline{2-5} 
                       & \multirow{4}{*}{PSO}    & 361K   & $D_{train}$& 9340.3 \\ \cline{3-5}                            &                         & 40K    & $D_{test}$& 1127.4 \\ \cline{3-5}  
                       &                         & 903K   & $D_{train}$& 25134.4 \\ \cline{3-5}                            &                         &   100K  & $D_{test}$& 2827.7 \\ \cline{2-5} 
                       & \multirow{4}{*}{GA}    & 361K   & $D_{train}$& 919.7 \\ \cline{3-5}                            &                         &  40K   & $D_{test}$& 75.6 \\ \cline{3-5} 
                       &                         & 903K   & $D_{train}$& 2161.0 \\ \cline{3-5}                            &                         &  100K   & $D_{test}$& 209.4\\ \hline 
\multirow{12}{*}{WAI Perf.} & \multirow{4}{*}{RS} & 361K   & $D_{train}$& 29.5 \\ \cline{4-5}                            &                         &   40K & $D_{test}$& 3.3 \\ \cline{3-5} 
                       &                         & 903K   & $D_{train}$& 67.2 \\ \cline{3-5}                            &                         &  100K   & $D_{test}$& 9.4 \\ \cline{2-5} 
                       & \multirow{4}{*}{PSO} & 361K   & $D_{train}$& 1099.8 \\ \cline{3-5}                            &                         &  40K   & $D_{test}$& 113.2 \\ \cline{3-5} 
                       &                         & 903K   & $D_{train}$& 3884.6 \\ \cline{3-5}                            &           &  100K   & $D_{test}$& 400.4 \\ \cline{2-5} 
                       & \multirow{4}{*}{GA}    & 361K   & $D_{train}$& 17.0 \\ \cline{3-5}                            &                         &   40K  & $D_{test}$& 1.2 \\ \cline{3-5} 
                       &                         & 903K   & $D_{train}$& 56.1 \\ \cline{3-5}                            &                         &  100K   & $D_{test}$& 12.1 \\ \hline 
\end{tabular}
\end{table}                         

As can be seen, the average counts of unique adversarial examples (i.e., Column Avg. \#AdvIn in Table~\ref{tab:RQ1}) are non-zeros. Actually, the foreseeable synthetic inputs that our approach produces to stress the conformity of models with physics-grounded sensitivity rules, have exposed physics inconsistencies regardless of the employed search algorithm.

\FrameSep.5em
\begin{framed}
\noindent
\textbf{Finding 1:} The designed physics-based adversarial testing successfully reveals the neural network’s inconsistencies against domain knowledge, assembling physics first principles and \textit{apriori} system design knowledge. 
\end{framed}

A comparison between the used search algorithms gives us the following ordered sequence: PSO, RS, then GA. The PSO-enabled generator succeeds in revealing the highest number of physics inconsistencies, and unexpectedly, the GA-enabled generator fails to even outperform the baseline, RS. To further compare the behaviors of these algorithms, we compute the average ratio of valid inputs generated by each algorithm, along with the average ratio of duplicate inputs that are discarded during the generation process, as shown in Table~\ref{tab:RQ2}. Indeed, the complexity of the encoded conditions could cause the search algorithm to stagnate in invalid regions without generating enough foreseeable inputs to test the model. In addition, a non-optimal exploitation/exploration trade off configuration would also cause the search algorithm to loop over the same regions of previously-triggered adversarial inputs, resulting in a high duplicate rate.

\begin{table}[ht]
\centering
\caption{Average ratio of valid test inputs and duplicated adversarial examples per SBST algorithm and per system}
\label{tab:RQ2}
\begin{tabular}{|l|l|l|l|}
\hline
\textbf{SYS} & \textbf{ALG} & \textbf{Avg. \%ValIn} & \textbf{Avg. \%DupIn}\\ \hline
\multirow{3}{*}{A-C Perf.}   & RS  & 31.40\%  & 0\%  \\ \cline{2-4} 
                       & PSO  & 44.35\%  & 31.61\% \\ \cline{2-4} 
                       & GA & 27.55\% &  10.47\% \\ \hline
\multirow{3}{*}{WAI Perf.} & RS  & 90.16\% & 0\% \\ \cline{2-4} 
                       & PSO  & 99.12\% & 81.56\%  \\ \cline{2-4} 
                       & GA & 82.45\% & 34.84\%  \\ \hline
\end{tabular}
\end{table}

As can be seen in Table\ref{tab:RQ2}, the average ratio of valid inputs per searching algorithm suggests the same sequence order we had in the comparison with respect to the counts of adversarial inputs. It is expected that the search approach that is able to remain in valid input regions longer, is more likely to find more adversarial examples. Although PSO was the most successful method in terms of valid and failed inputs, it also produced the highest ratio of duplicate inputs. We find this to be a strong indicator of over-exploitation, as PSO might have been heavily relying on the most-fitted previous solutions, which kept it rolling over the same solutions indefinitely. These observed differences in the search behaviors of the evaluated algorithms indicate that they produce inputs of varying diversity. Hence in the following research questions, we consider all three search algorithms in assessing the regularization based on physics domain knowledge, both in terms of fixing the physics inconsistencies and improving the prediction errors. We intend to explore the effects of AXs source generators on model improvements.

\FrameSep.5em
\begin{framed}
\noindent
\textbf{Finding 2:} our PSO-enabled testing approach outperforms GA-enabled version and RS baseline in regards to the number of successfully-exposed physics inconsistencies of the trained neural networks. 
\end{framed}

\begin{figure}[h]
\centering
\includegraphics[scale=0.34]{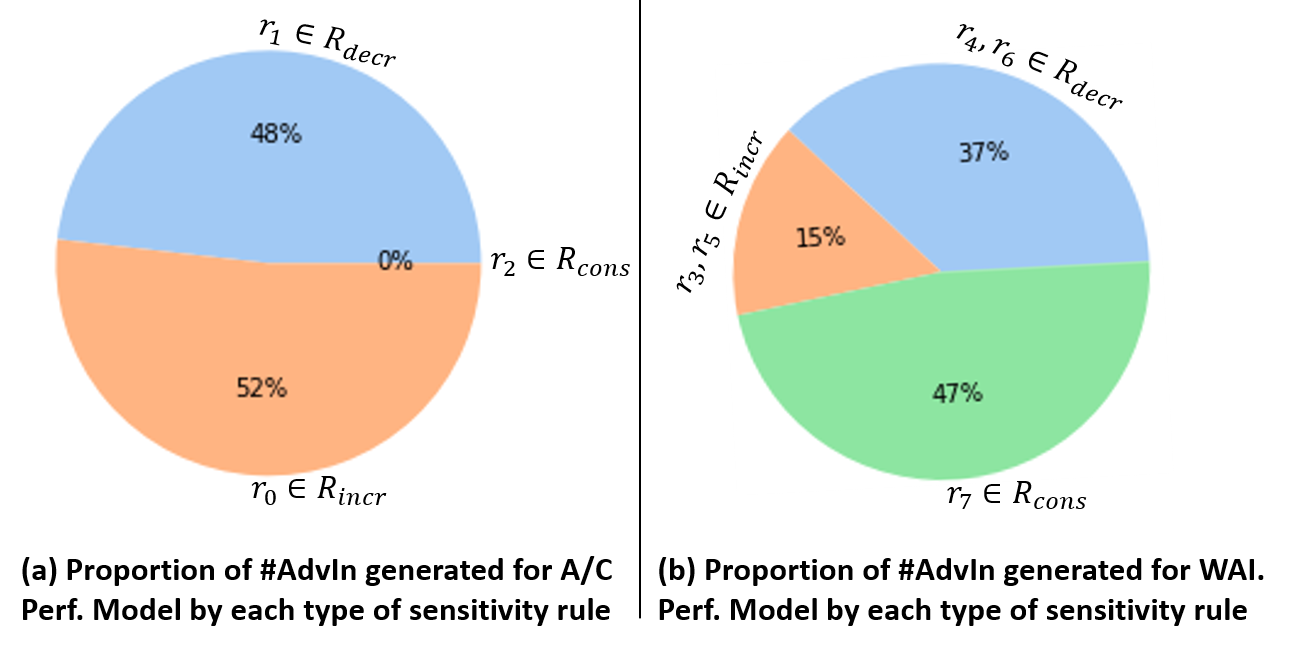}
\caption{Proportion of \#AdvIn generated by each Type of Sensitivity Rule}
\label{fig:propos_AdvIn}
\end{figure}

Considering the smoothness of the simulated systems' dynamics, a weight decay regularization was applied to system performance models to promote the learning of smooth mapping functions and penalize unnecessarily large response changes. Nevertheless, Figure~\ref{fig:propos_AdvIn} shows the proportion of the adversarial inputs generated by our approach using each type of sensitivity rule for (a) the A/C performance model, which is resilient to invariance tests, and (b) the WAI performance model, which is highly sensitive to input perturbations with a large proportion of failed invariance tests. This observed difference in proneness to injected input noises can be attributed to the complexity of the system and the size of the datasets. The limited number of steady-state flights with the underlying WAI system deployed prevents the NN-based simulator from learning numerically stable states, leaving it susceptible to invariant adversarial examples. Thereby, our physics-informed adversarial training relies on failed invariance tests to increase the numerical stability of the neural network over subsequent fine-tuning steps. Moreover, Figure~\ref{fig:propos_AdvIn} shows that the A/C performance model (a) produces almost equally-distributed directional expectation tests, precisely the increasing and decreasing output rules. Alternatively, the WAI performance model (b) generates a higher percentage of AXs based on decreasing output rules as compared to the increasing output counterparts. This observed difference can be explained by the proximity to normal behaviors. On the one hand, the WAI system should maintain the skin at warm or hot temperatures as necessary, depending on the ambient conditions. On the other hand, the decreasing-output sensitivity rules define the input perturbations causing the outputs to decrease. Thereby, these sensitivity rules derive synthetic flight datapoints, having skin temperatures that are expected to be lower than the original flight test inputs. Hence, the resulting decreased-output datapoints are capable of simulating rare and extreme operating conditions to challenge the original performance model trained on left-skewed data distribution (i.e., skin temperatures tend to be relatively high compared to the foreseeable range).

\textbf{Domain Experts Feedback. }Aircraft development engineers perceive the potential value of our proposed adversarial testing approach in the ML-based simulator development. They affirm that the sensitivity analysis represents a main step in the system modelling to validate their expectations through what-if scenarios on how the target variables should be affected based on changes in the input variables. The what-if question would be like ``what would happen to the quantity of interest $q$ if the input variable $var_1$ went up by $1\%$". Therefore, they consider our proposed SBST approach as kind of a large-scale automation for searching over all the possible what-if scenarios and reporting the ones triggering unanticipated behaviors. Nonetheless, our system engineering stakeholders emphasize the challenges of defining the application scope of such physics-grounded sensitivity rules in regards to more sophisticated systems. Indeed, further restrictions on the local subspace of perturbed inputs would likely take place when combined variations of system inputs as well as relatively high magnitude changes provoke non-negligible higher order effects on the target output.

\textbf{\textit{RQ2. How well does the physics-informed regularization fix the physics inconsistencies?}}

\textbf{Motivation. }The goal is to assess the usefulness of our proposed physics-informed regularization in immuning the DNN against physics-based adversarial examples. In other words, we want to assess how well the regularized model learns the underlying physics sensitivity rules. 

\textbf{Method. }The adversarial examples detected by each search algorithm during the second test session, when a larger number of generations is involved, are loaded onto the training datasets. Then, we perform physics-informed adversarial training that combines the conventional data loss with our proposed regularization cost for the adversarial examples revealed in the training datasets. Thus, we re-run the adversarial detection process on the adversarial examples exposed for its corresponding validation dataset. We aim to check whether the detected physics inconsistencies in the validation data stay or disappear. This allows us to assess the effectiveness of our physics-informed adversarial training phase in fine-tuning the neural network to the foreknown relationships between the input and output variables. As discussed in the previous research question, we keep all three search algorithms in the analysis to compare their resulting regularization improvements, separately.

\textbf{Results.} Table~\ref{tab:RQ3} compares the number of physics inconsistencies (\#AdvIn) revealed before and after the physics-informed adversarial learning, as well as the estimated improvement ratios (\%Improv\_AdvIn) for each supported search algorithm.

\begin{table}[ht]
\centering
\caption{comparison between \#adversarials before and after fine-tuning fix}
\label{tab:RQ3}
\begin{tabular}{|l|l|l|l|l|l|}
\hline
\textbf{SYS} & \textbf{G} & \textbf{pre-\#AdvIn} & \textbf{post-\#AdvIn} & \textbf{\%Improv\_AdvIn} \\ 
\hline
\multirow{3}{*}{A-C Perf.}   & RS  & 5267  & 1012   & 80.78\% \\ \cline{2-5} 
                             & PSO  & 39551  & 5747	  & 85.46\%        \\ \cline{2-5} 
                             & GA & 2850 & 636  & 77.68\%        \\ \hline
\multirow{3}{*}{WAI Perf.}   & RS  & 509  & 0   & 100\% \\ \cline{2-5} 
                       & PSO  & 20545	 & 18   &  99.91\%       \\ \cline{2-5} 
                       & GA &  459  & 4   &  99.12\%       \\ \hline
\end{tabular}
\end{table}

As illustrated by the improvement ratios in Table~\ref{tab:RQ3}, the regularized neural network has triggered less physics inconsistencies in regards to the validation datasets for both of the studied performance models and the three implemented search algorithms. Therefore, the inclusion of physics-informed regularization costs improves the neural networks' propensity to be consistent with the underlying physics sensitivity rules. 

Besides, search algorithms with higher counts of exposed AXs for the base neural network have achieved higher improvement ratios (\%Improv\_AdvIn) of the physics inconsistencies. Indeed, sorting the search algorithms by the column of Table~\ref{tab:RQ3}, \#pre-AdvIn (numbers of exposed AXs for base NNs), returns PSO, RS, followed by GA, which achieved on average, respectively, $92.69\%$, $90.4\%$, and $88.4\%$. The ranks of search algorithms obtained by \%Improv\_AdvIn are in agreement with their ranks by \#AdvIn (i.e., number of exposed AXs) on both training and testing datasets, as shown in Table~\ref{tab:RQ1}. Given the statistical grounds and data-driven nature of the proposed physics-informed regularization cost, this result is expected. In fact, our physics-guided adversarial training incorporates soft penalty terms for violations of physics-based sensitivities to the loss function. Then, it proceeds with data-driven repairs of the neural network using the semi-supervised AX datasets detected on the training datasets. Thus, the search algorithms that yield larger AX datasets during the adversarial testing, provide more physics inconsistencies to teach the model the specified physics-grounded sensitivities over fine-tuning steps with the proposed composite loss function. Therefore, their corresponding regularized neural networks are unlikely to behave inconsistently with the foreknown sensitivities, even on the synthetic inputs crafted from test datasets that might differ from the ones built using the training examples.

In addition to the influence of the AX data size, we also noticed that the initial problem size measured by the dimensions of the original data, affects the effectiveness of our physics-informed adversarial training. In Table~\ref{tab:RQ3}, we can observe that the improvement ratios obtained for the WAI performance model are in-between $99\%$-$100\%$, in comparison with the achieved improvements on the A/C performance model, ranging from $77\%$ to $85\%$. This distance in the number of fixed physics inconsistencies between the two studied simulation problems could be explained by the absolute number of revealed AXs and the size of the original datasets, as demonstrated in Table~\ref{tab:size-dim-dataset}. As can be seen, the WAI performance data has less original datapoints ($90$) and more feature dimensions ($10$), which results in a low input space coverage and less diversity in the revealed AXs, while the A/C performance data has more datapoints ($1334$) and less feature dimensions ($5$), yielding a higher input space coverage and more diversity in the revealed AXs.

\FrameSep.5em
\begin{framed}
\noindent
\textbf{Finding 3:} The physics-informed regularization improves the quality of the learned model’s mappings towards more consistency with the physics domain knowledge, but the achieved improvements depend on the source of physics-based AXs and the genuine input space coverage.
\end{framed}

\textbf{Domain Experts Feedback. }Regarding the data-driven corrections of the ML model's physics consistency, system engineering experts introduce us the possible noises in the test flight sensors data that are leveraged to train and test the ML models. In fact, earlier versions of our approach did not include an expert-defined tolerable deviation ($tol_j$) that has been used for both of the follow-up adversarial input test and physics-informed regularization cost. Formerly, our designed adversarial ML approach reports consistently a high number of revealed adversarial inputs due to a slightly deviation of their outputs in the undesired direction. Moreover, the rigid physics-informed regularization cost without tolerance parameter was kind of pushing the fine-tuned neural model towards over-respecting the designed theory-grounded rules on the sensors data collected during monitored flight tests.

\textbf{\textit{RQ3. Does the physics-informed loss improves or degrades the performance of the DNN?}}

\textbf{Motivation. }The purpose of this research question is to determine whether physics-informed loss affects the performance of the fine-tuned models indirectly.

\textbf{Method. }We took the fine-tuned models from the physics-informed adversarial training, and we computed their predictive performance, i.e., the root mean squared error (RMSE), on the original data. 

\textbf{Results. }The RMSE and \%Change\_RMSE values, shown in Table~\ref{tab:RQ4}, are obtained from the evaluation of neural networks on the validation datasets after and before performing the fine-tuning sessions on semi-supervised datasets that include the adversarial examples found on the training examples. 

\begin{table}[ht]
\centering
\caption{Comparison between RMSE after and before fine-tuning fix}
\label{tab:RQ4}
\begin{tabular}{|l|l|l|l|l|l|}
\hline
\textbf{SYS} & \textbf{TRG}   & \textbf{pre-RMSE}  & \textbf{ALG} & \textbf{post-RMSE} & \shortstack{\textbf{\%Chg} \\ \textbf{RMSE}}\\\hline
\multirow{3}{*}{A-C Perf.}   
& \multirow{3}{*}{$\alpha$} 
& \multirow{3}{*}{0.498°}  
& RS& 0.497°& \textcolor{dkgreen}{-0.20}\\ \cline{4-6} 
&  &  & PSO & 0.996°& \textcolor{red}{+100}\\ \cline{4-6} 
&  &  & GA& 0.444°& \textcolor{dkgreen}{-10.84}\\ \hline
\multirow{6}{*}{WAI Perf.} & 
\multirow{3}{*}{$T_{skin}^a$}& 
\multirow{3}{*}{4.088°C} 
& RS & 4.729°C& \textcolor{red}{+15.68}\\ \cline{4-6} 
&  &  & PSO & 4.422°C& \textcolor{red}{+8.17}\\ \cline{4-6} 
&  &  & GA & 3.979°C& \textcolor{dkgreen}{-2.67}\\ \cline{2-6} 
& \multirow{3}{*}{$T_{skin}^b$} & 
\multirow{3}{*}{7.524°C} 
& RS & 7.921°C& \textcolor{red}{+5.28}\\ \cline{4-6} 
&  &  & PSO& 6.826°C& \textcolor{dkgreen}{-9.28}\\ \cline{4-6} 
&  &  & GA& 7.163°C& \textcolor{dkgreen}{-4.80}\\ \hline
\end{tabular}
\end{table}

Based on Table~\ref{tab:RQ4}, we found that random sampler (RS) with no searching capabilities caused the RMSE to either stall or increase in all the assessment experiences. In contrast, GA-enabled input generation leads to only negative \%Change\_RMSEs, which means that the post-fix RMSEs have been successfully decreased. PSO, which was the most effective in exposing the physics-based AXs with higher \#AdvIn, often degrades post-regularization RMSEs (i.e., yielding positive \%Change\_RMSE) at the cost of reducing the inconsistencies with respect to the physics-grounded sensitivity rules. As demonstrated by the study case on A/C performance, PSO found much higher AXs than other competitors (see Table~\ref{tab:RQ1}) and achieved better physics consistency improvements (see Table~\ref{tab:RQ3}). The post-fix RMSE, however, was doubled, with \%Change\_RMSE equal to $100\%$.

In our experimentation, GA-enabled adversarial testing has identified few but useful adversarial inputs that improve post-regularization RMSEs, contrary to PSO and RS, which were less effective in identifying adversarial examples that could positively affect the predictive performance of regularized neural networks after fine-tuning. While aircraft performance studies often involve limited-size datasets of expensive flight tests, they are intended to train simulation models of the system's expected behavior under all foreseeable operating conditions. Hence, an increase in the estimated errors on validation data will be acceptable within a certain system-dependent range validated by domain experts to ensure that the statistically-derived mappings of the model are reliable and fairly consistent with the physics domain knowledge. Using the same A/C performance model as an example, our collaborators contend that a regularized model with a 1° of average error and high consistency with the specified physics sensitivities over the flight envelope can be more useful in different engineering applications than an A/C performance model with low physics consistency and a 0.5° of average error.

\FrameSep.5em
\begin{framed}
\noindent
\textbf{Finding 4:} A physics-informed adversarial training's effect on the predictive performance heavily depends on the size and quality of the revealed adversarial input data. According to our experiments, these AX criteria can be controlled by selecting the appropriate search algorithm. Specifically, the GA-enabled generator reveals a few physics-based AXs that improve the prediction, while the PSO counterpart reveals many AXs, but at the risk of degrading the original predictions.
\end{framed}

\textbf{Domain Experts Feedback. }Regarding the probable degradation induced by the physics-informed adversarial training, aircraft system engineering experts highlight the importance that data-driven models statistically infer input-output mappings, which are thoroughly consistent with the physics-grounded sensitivity rules. Thus, they suggest that the physics-guided adversarial approach should allow the user to manually balance out the tradeoff between the data loss and the physics-informed loss during the physics-guided adversarial training, depending on the system modelling use case. In response to that, we design a weight parameter, $\beta$, to control the regularization cost similarly to the traditional norm penalties, while keeping the dynamic calibration of $\lambda$ that we added to overcome the differences in the losses magnitudes. Thus, we will have $\lambda = \beta\times\lambda_{dynamic}$, where the $\lambda_{dynamic}$ is the actual pre-computed lambda value and the $\beta \geq 1$ is the weight of regularization cost, by default equals to $1$. For some use cases, system engineers can set up higher $\beta$ values to prioritize the model's physics consistency within the input space over further improvements on the RMSE. 
\section{Conclusion}
\label{sec:conclusion}
The present paper proposes a physics-guided adversarial machine learning approach that assembles: (i) a physics-guided adversarial testing method that successfully exposes physics inconsistencies in ML-based A/C systems performance models; (ii) a physics-informed adversarial training approach that promotes learning input-output mappings, satisfying the desired level of consistency with physics domain knowledge. A physics inconsistency in the input space could be expected owing to the complexity of the simulated system dynamics and the rarity of the flight test data. Our physics-based adversarial testing applies search algorithms to conduct worst-case analysis on the model’s inconsistencies and provide insights on their prevalence in the input space. Our subsequent physics-informed regularization always improves the physics consistency of the model, but we observed that a high density of exposed AXs might degrade its predictive performance. In the future work, we focus on empowering the aircraft engineers with more control of the trade-off between physics consistency and prediction errors on the validation datasets. Hence, we aim to connect tightly the adversarial search problem to the model regularization results; so the users can tune our approach on their models to the best consistency-error trade-off for the underlying aircraft engineering study cases. Moreover, our physics-guided adversarial ML approach can have profound implications for statistical model engineering. Indeed, our physics-informed loss can also steer the users towards designing neural networks with structure and hyperparameters that are more suitable for solving the consistency-error trade-off. As long as the physics-based sensitivity rules can be specified, a trustworthy aircraft system performance model should not only have a low prediction error, but it should also be consistent with the physics domain knowledge.
\section*{Acknowledgment}
This work is supported by the DEEL project CRDPJ 537462-18 funded by the National Science and Engineering Research Council of Canada (NSERC) and the Consortium for Research and Innovation in Aerospace in Québec (CRIAQ), together with its industrial partners Thales Canada inc, Bell Textron Canada Limited, CAE inc and Bombardier inc.
\balance
\bibliography{main}{}

\begin{thebibliography}{10}
\providecommand{\url}[1]{#1}
\csname url@samestyle\endcsname
\providecommand{\newblock}{\relax}
\providecommand{\bibinfo}[2]{#2}
\providecommand{\BIBentrySTDinterwordspacing}{\spaceskip=0pt\relax}
\providecommand{\BIBentryALTinterwordstretchfactor}{4}
\providecommand{\BIBentryALTinterwordspacing}{\spaceskip=\fontdimen2\font plus
\BIBentryALTinterwordstretchfactor\fontdimen3\font minus
  \fontdimen4\font\relax}
\providecommand{\BIBforeignlanguage}[2]{{%
\expandafter\ifx\csname l@#1\endcsname\relax
\typeout{** WARNING: IEEEtran.bst: No hyphenation pattern has been}%
\typeout{** loaded for the language `#1'. Using the pattern for}%
\typeout{** the default language instead.}%
\else
\language=\csname l@#1\endcsname
\fi
#2}}
\providecommand{\BIBdecl}{\relax}
\BIBdecl

\bibitem{marini2002verification}
M.~Marini, R.~Paoli, F.~Grasso, J.~Periaux, and J.-A. Desideri, ``Verification
  and validation in computational fluid dynamics: the flownet database
  experience,'' \emph{JSME International Journal Series B Fluids and Thermal
  Engineering}, vol.~45, no.~1, pp. 15--22, 2002.

\bibitem{roache1998verification}
P.~J. Roache, \emph{Verification and validation in computational science and
  engineering}.\hskip 1em plus 0.5em minus 0.4em\relax Hermosa Albuquerque, NM,
  1998, vol. 895.

\bibitem{ioannou2018structural}
Y.~A. Ioannou, ``Structural priors in deep neural networks,'' Ph.D.
  dissertation, University of Cambridge, 2018.

\bibitem{ren2018learning}
H.~Ren, R.~Stewart, J.~Song, V.~Kuleshov, and S.~Ermon, ``Learning with weak
  supervision from physics and data-driven constraints,'' \emph{AI Magazine},
  vol.~39, no.~1, pp. 27--38, 2018.

\bibitem{stewart2017label}
R.~Stewart and S.~Ermon, ``Label-free supervision of neural networks with
  physics and domain knowledge,'' in \emph{Thirty-First AAAI Conference on
  Artificial Intelligence}, 2017.

\bibitem{muralidhar2019physics}
N.~Muralidhar, J.~Bu, Z.~Cao, L.~He, N.~Ramakrishnan, D.~Tafti, and
  A.~Karpatne, ``Physics-guided deep learning for drag force prediction in
  dense fluid-particulate systems,'' \emph{Big Data}, vol.~8, no.~5, pp.
  431--449, 2020.

\bibitem{wang2020recent}
X.~Wang, Y.~Zhao, and F.~Pourpanah, ``Recent advances in deep learning,'' 2020.

\bibitem{biggio2018wild}
B.~Biggio and F.~Roli, ``Wild patterns: Ten years after the rise of adversarial
  machine learning half-day tutorial,'' in \emph{25th ACM Conference on
  Computer and Communications Security, CCS 2018}.\hskip 1em plus 0.5em minus
  0.4em\relax Association for Computing Machinery, 2018, pp. 2154--2156.

\bibitem{szegedy2013intriguing}
C.~Szegedy, W.~Zaremba, I.~Sutskever, J.~Bruna, D.~Erhan, I.~Goodfellow, and
  R.~Fergus, ``Intriguing properties of neural networks,'' \emph{arXiv preprint
  arXiv:1312.6199}, 2013.

\bibitem{goodfellow2014explaining}
I.~J. Goodfellow, J.~Shlens, and C.~Szegedy, ``Explaining and harnessing
  adversarial examples,'' \emph{arXiv preprint arXiv:1412.6572}, 2014.

\bibitem{nguyen2018adversarial}
A.~T. Nguyen and E.~Raff, ``Adversarial attacks, regression, and numerical
  stability regularization,'' in \emph{AAAI-19 Workshop on Engineering
  Dependable and Secure Machine Learning Systems}.\hskip 1em plus 0.5em minus
  0.4em\relax AAAI, 2019.

\bibitem{mcminn2004search}
P.~McMinn, ``Search-based software test data generation: a survey,''
  \emph{Software testing, Verification and reliability}, vol.~14, no.~2, pp.
  105--156, 2004.

\bibitem{deepevolution}
H.~Ben~Braiek and F.~Khomh, ``Deepevolution: A search-based testing approach
  for deep neural networks,'' in \emph{2019 IEEE International Conference on
  Software Maintenance and Evolution}.\hskip 1em plus 0.5em minus 0.4em\relax
  IEEE, 2019.

\bibitem{bhambri2019survey}
S.~Bhambri, S.~Muku, A.~Tulasi, and A.~B. Buduru, ``A survey of black-box
  adversarial attacks on computer vision models,'' \emph{arXiv preprint
  arXiv:1912.01667}, 2019.

\bibitem{mosli2019they}
R.~Mosli, M.~Wright, B.~Yuan, and Y.~Pan, ``They might not be giants: Crafting
  black-box adversarial examples with fewer queries using particle swarm
  optimization,'' in \emph{Computer Security -- ESORICS 2020}.\hskip 1em plus
  0.5em minus 0.4em\relax Springer International Publishing, 2020, pp.
  439--459.

\bibitem{alzantot2019genattack}
M.~Alzantot, Y.~Sharma, S.~Chakraborty, H.~Zhang, C.-J. Hsieh, and M.~B.
  Srivastava, ``Genattack: Practical black-box attacks with gradient-free
  optimization,'' in \emph{Proceedings of the Genetic and Evolutionary
  Computation Conference}, 2019, pp. 1111--1119.

\bibitem{chen2019poba}
J.~Chen, M.~Su, S.~Shen, H.~Xiong, and H.~Zheng, ``Poba-ga: Perturbation
  optimized black-box adversarial attacks via genetic algorithm,''
  \emph{Computers \& Security}, vol.~85, pp. 89--106, 2019.

\bibitem{wang2019natural}
X.~Wang, H.~Jin, and K.~He, ``Natural language adversarial attacks and defenses
  in word level,'' \emph{arXiv preprint arXiv:1909.06723}, 2019.

\bibitem{zang2020word}
Y.~Zang, F.~Qi, C.~Yang, Z.~Liu, M.~Zhang, Q.~Liu, and M.~Sun, ``Word-level
  textual adversarial attacking as combinatorial optimization,'' in
  \emph{Proceedings of the 58th Annual Meeting of the Association for
  Computational Linguistics}, 2020, pp. 6066--6080.

\bibitem{du2020hybrid}
X.~Du, J.~Yu, Z.~Yi, S.~Li, J.~Ma, Y.~Tan, and Q.~Wu, ``A hybrid adversarial
  attack for different application scenarios,'' \emph{Applied Sciences},
  vol.~10, no.~10, p. 3559, 2020.

\bibitem{du2020sirenattack}
T.~Du, S.~Ji, J.~Li, Q.~Gu, T.~Wang, and R.~Beyah, ``Sirenattack: Generating
  adversarial audio for end-to-end acoustic systems,'' in \emph{Proceedings of
  the 15th ACM Asia Conference on Computer and Communications Security}, 2020,
  pp. 357--369.

\bibitem{PSO}
R.~Eberhart and J.~Kennedy, ``A new optimizer using particle swarm theory,'' in
  \emph{Micro Machine and Human Science, 1995. MHS'95., Proceedings of the
  Sixth International Symposium on}.\hskip 1em plus 0.5em minus 0.4em\relax
  IEEE, 1995, pp. 39--43.

\bibitem{GA}
J.~H. Holland, ``Genetic algorithms,'' \emph{Scientific american}, vol. 267,
  no.~1, pp. 66--73, 1992.

\bibitem{picek2010comparison}
S.~Picek and M.~Golub, ``Comparison of a crossover operator in binary-coded
  genetic algorithms,'' \emph{WSEAS transactions on computers}, vol.~9, no.~9,
  pp. 1064--1073, 2010.

\bibitem{karpatne2017theory}
A.~Karpatne, G.~Atluri, J.~H. Faghmous, M.~Steinbach, A.~Banerjee, A.~Ganguly,
  S.~Shekhar, N.~Samatova, and V.~Kumar, ``Theory-guided data science: A new
  paradigm for scientific discovery from data,'' \emph{IEEE Transactions on
  knowledge and data engineering}, vol.~29, no.~10, pp. 2318--2331, 2017.

\bibitem{ling2016reynolds}
J.~Ling, A.~Kurzawski, and J.~Templeton, ``Reynolds averaged turbulence
  modelling using deep neural networks with embedded invariance,''
  \emph{Journal of Fluid Mechanics}, vol. 807, pp. 155--166, 2016.

\bibitem{seo2019differentiable}
S.~Seo and Y.~Liu, ``Differentiable physics-informed graph networks,'' in
  \emph{ICLR-19 Workshop}, 2019.

\bibitem{leibo2017view}
J.~Z. Leibo, Q.~Liao, F.~Anselmi, W.~A. Freiwald, and T.~Poggio,
  ``View-tolerant face recognition and hebbian learning imply mirror-symmetric
  neural tuning to head orientation,'' \emph{Current Biology}, vol.~27, no.~1,
  pp. 62--67, 2017.

\bibitem{ho2002simple}
Y.-C. Ho and D.~L. Pepyne, ``Simple explanation of the no-free-lunch theorem
  and its implications,'' \emph{Journal of optimization theory and
  applications}, vol. 115, no.~3, pp. 549--570, 2002.

\bibitem{joshi2020parameter}
S.~K. Joshi and J.~C. Bansal, ``Parameter tuning for meta-heuristics,''
  \emph{Knowledge-Based Systems}, vol. 189, p. 105094, 2020.

\bibitem{hinge_loss}
L.~Rosasco, E.~De~Vito, A.~Caponnetto, M.~Piana, and A.~Verri, ``Are loss
  functions all the same?'' \emph{Neural computation}, vol.~16, no.~5, pp.
  1063--1076, 2004.

\bibitem{cfr}
\BIBentryALTinterwordspacing
``14 cfr § 33.68 - induction system icing.'' November 2014. [Online].
  Available: \url{https://www.law.cornell.edu/cfr/text/14/33.68}
\BIBentrySTDinterwordspacing

\bibitem{eshelby2000aircraft}
M.~Eshelby, \emph{Aircraft performance: Theory and practice}.\hskip 1em plus
  0.5em minus 0.4em\relax American Institute of Aeronautics and Astronautics,
  Inc., 2000.

\bibitem{hahn2012heat}
D.~W. Hahn and M.~N. {\"O}zisik, \emph{Heat conduction}.\hskip 1em plus 0.5em
  minus 0.4em\relax John Wiley \& Sons, 2012.

\bibitem{incropera1996fundamentals}
F.~P. Incropera, D.~P. DeWitt, T.~L. Bergman, A.~S. Lavine \emph{et~al.},
  \emph{Fundamentals of heat and mass transfer}.\hskip 1em plus 0.5em minus
  0.4em\relax Wiley New York, 1996, vol.~6.

\bibitem{design_principles}
S.~H. Hasanpour, M.~Rouhani, M.~Fayyaz, M.~Sabokrou, and E.~Adeli, ``Towards
  principled design of deep convolutional networks: Introducing simpnet,''
  \emph{arXiv preprint arXiv:1802.06205}, 2018.

\bibitem{nvidia}
\BIBentryALTinterwordspacing
V.~Sarge, M.~Andersch, L.~Fabel, P.~Micikevicius, and J.~Tran, ``Tips for
  optimizing gpu performance using tensor cores,'' June 2019. [Online].
  Available:
  \url{https://developer.nvidia.com/blog/optimizing-gpu-performance-tensor-cores/}
\BIBentrySTDinterwordspacing

\bibitem{paszke2019pytorch}
A.~Paszke, S.~Gross, F.~Massa, A.~Lerer, J.~Bradbury, G.~Chanan, T.~Killeen,
  Z.~Lin, N.~Gimelshein, L.~Antiga \emph{et~al.}, ``Pytorch: An imperative
  style, high-performance deep learning library,'' in \emph{Advances in neural
  information processing systems}, 2019, pp. 8026--8037.

\end{thebibliography}
\bibliographystyle{IEEEtran}
\end{document}